\documentclass[twoside]{article}
\usepackage{textcomp}
\usepackage[accepted]{aistats2021}

\usepackage{capt-of}
\usepackage{booktabs} 

\usepackage{hyperref}

\usepackage[round]{natbib}

\usepackage[pdftex]{graphicx}

\usepackage{amsmath}
\usepackage{amsthm}
\usepackage{amssymb}
\usepackage{enumitem}
\usepackage{float}
\usepackage{array}
\usepackage{multirow}

\usepackage{tikz,forest}
\usetikzlibrary{arrows.meta}

\usepackage{algorithm}
\usepackage[noend]{algorithmic}

\usepackage{forloop}
\newcounter{ct}

\newcommand{\Loss}{\mathcal{L}}
\newcommand{\M}{\mathcal{P}}
\newcommand{\Q}{\mathcal{Q}}
\newcommand{\V}{\mathcal{V}}

\newcommand{\W}{\mathcal{W}}
\newcommand{\X}{\mathcal{X}}
\newcommand{\A}{\mathcal{A}}
\newcommand{\State}{\mathcal{S}}
\newcommand{\Rew}{\mathcal{R}}
\newcommand{\Rmax}{R_{\text{max}}}

\newcommand{\density}[2]{d^{#2}_{#1,\gamma}}
\newcommand{\densityt}[3]{d^{#2}_{#1,#3}}
\newcommand{\val}[2]{V^{#2}_{#1}}

\newcommand{\Hh}{\mathcal{H}}

\newcommand{\R}{\mathbb{R}}

\usepackage{color}

\newcommand*\xbar[1]{%
   \hbox{%
     \vbox{%
       \hrule height 0.5pt 
       \kern0.5ex
       \hbox{%
         \kern-0.1em
         \ensuremath{#1}%
         \kern-0.1em
       }%
     }%
   }%
} 
\graphicspath{{figures/}}

\theoremstyle{plain}
\newtheorem{thm}{Theorem}[section]
\newtheorem{lem}[thm]{Lemma}
\newtheorem{prop}[thm]{Proposition}
\newtheorem{cor}[thm]{Corollary}
\newtheorem{assumption}{Assumption}

\newtheorem{rem}[thm]{Remark}

\theoremstyle{definition}
\newtheorem{defn}{Definition}[section]

\newtheorem{exmp}{Example}[section]

\theoremstyle{remark}

\newcommand*{\starnr}{\stepcounter{equation}\tag{\theequation}}
\makeatletter
\newenvironment{Salign}
  {\start@align\@ne\st@rredtrue\m@ne}
  {\starnr\endalign}
\makeatother

\begin{document}

%

%

\twocolumn[

\aistatstitle{Minimax Model Learning}

\aistatsauthor{ Cameron Voloshin \And Nan Jiang \And  Yisong Yue }

\aistatsaddress{ Caltech \And  UIUC \And Caltech } ]

\begin{abstract}
We  present  a  novel  off-policy  loss  function  for learning a transition model in model-based reinforcement learning.  Notably, our loss is derived from the off-policy policy evaluation objective with  an  emphasis  on  correcting distribution shift. 
Compared to previous model-based techniques, our approach allows for greater robustness under model mis-specification or distribution shift  induced  by  learning/evaluating policies that are distinct from the data-generating policy. We provide a theoretical analysis and show empirical improvements over existing model-based off-policy evaluation methods. We provide further analysis showing our loss can be used for off-policy optimization (OPO) and demonstrate its integration with more recent improvements in OPO.
\end{abstract}
\section{Introduction}




We study the problem of learning a transition model in a batch, off-policy reinforcement learning (RL) setting, i.e., of learning a function $P(s'|s,a)$ 
from a pre-collected dataset $D = \{(s_i,a_i,s'_i)\}_{i=1}^n$ without further access to the environment. 
Contemporary approaches to model learning focus primarily on improving the performance of models learned through maximum likelihood estimation (MLE)  \citep[]{Sutton1990Dyna, DeisenrothPILCO2011, kurutach2018metrpo, Clavera2018MBMPO, Chua2018PETS, Luo2019SLBO}. The goal of MLE is to pick the model within some model class $\M$ that is most consistent with the observed data or, equivalently, most likely to have generated the data. This is done by minimizing negative log-loss (minimizing the KL divergence) summarized as follows: 
\begin{align}
    \widehat{P}_{\text{MLE}} &= \arg\min_{P \in \M} \frac{1}{n}\sum_{(s_i,a_i,s'_i) \in D} -\log(P(s_i'|s_i,a_i)). \label{eqn:mle}
\end{align}
A key limitation of MLE is that it focuses on picking a good model under the data distribution while ignoring how the model is actually used. 

In an RL context, a model can be used to either learn a policy (policy learning/optimization) or evaluate some given policy (policy evaluation), without having to collect more data from the true environment. We call this actual objective the ``decision problem." Interacting with the environment to solve the decision problem can be difficult, expensive and dangerous, whereas a model learned from batch data circumvents these issues. Since MLE \eqref{eqn:mle} does not optimize over the distribution of states induced by the policy from the decision problem, it thus does not prioritize solving the decision problem. 
Notable previous works that incorporate the decision problem into the model learning objective are Value-Aware Model Learning (VAML) and its variants 
\citep[]{farahmand2017VAML, farahmand2018IVAML, abachi2020PAML}. 
%
These methods, however, still define their losses w.r.t.~the data distribution as in MLE, and ignore the \emph{distribution shift} from the pre-collected data to the policy-induced distribution.

In contrast, we directly focus on requiring the model to perform well under unknown distributions instead of the data distribution. 
In other words, we are particularly interested in developing approaches that directly model the batch (offline) learning setting.
%
As such, we ask: \textit{``From only pre-collected data, is there a model learning approach that naturally controls the decision problem error?"}

In this paper, we present a new loss function for model learning that: (1) only relies on batch or offline data;
 (2) takes into account the distribution shift effects; and (3) directly relates to the performance metrics for off-policy evaluation and learning under certain realizability assumptions.  
The design of our loss is inspired by recent advances in model-free off-policy evaluation \citep[e.g.,][]{liu2018breaking, uehara2019minimax},
which we build upon to develop our approach.
\section{Preliminaries}

We adopt the infinite-horizon discounted MDP framework specified by a tuple $(\State, \A, P, \Rew, \gamma)$, where $\State$ is the state space, $\A$ is the
action space, $P : \State \times \A \to \Delta(\State)$ is the transition function,
$\Rew : \State \times \A \to \Delta([-\Rmax, \Rmax])$ is the reward function, and
$\gamma \in [0, 1)$ is the discount factor. Let $\X \equiv \State \times \A$. Given an MDP,
a (stochastic) policy $\pi: \State \to \Delta(\A)$ and a starting state
distribution $d_0 \in \Delta(\State)$ together determine a distribution
over trajectories of the form $s_0, a_0, r_0, s_1, a_1, r_1, \ldots,$ where
$s_0 \sim d_0, a_t \sim \pi(s_t), r_t \sim \Rew(s_t, a_t)$, and $s_{t+1} \sim P(s_t, a_t)$ for $t \geq 0$. 
The performance of policy $\pi$  is given by:
\begin{equation}\label{def:JV}
    J(\pi, P) \equiv E_{s \sim d_0}[\val{\pi}{P}(s)],
\end{equation} 
where, by the Bellman Equation, 
\begin{small}
\begin{equation} \label{def:Bellman}
\val{\pi}{P}(s) \equiv E_{a \sim \pi(\cdot|s)}[E_{r \sim \mathcal{R}(\cdot|s,a)}[r] + \gamma E_{\tilde{s} \sim P(\cdot|s,a)}[\val{\pi}{P}(\tilde{s})]].
\end{equation} 
\end{small}
\hspace{-3pt}A useful equivalent measure of performance is:
\begin{equation}\label{def:JW}
    J(\pi, P) = E_{(s,a) \sim \density{\pi}{P}}[E_{ r \sim \mathcal{R}(\cdot| s,a)}[r]],
\end{equation}
where $\density{\pi}{P}(s,a) \equiv  \sum_{t=0}^\infty \gamma^t \densityt{\pi}{P}{t}(s,a)$ is the (discounted) distribution of state-action pairs induced by running $\pi$ in $P$ and $\densityt{\pi}{P}{t} \in \Delta(\X)$ is the distribution of $(s_t, a_t)$ induced by running $\pi$ under $P$. The first term in $\density{\pi}{P}$ is $\densityt{\pi}{P}{0} = d_0$.  $\densityt{\pi}{P}{t}$ has a recursive definition that we use in Section \ref{sec:mml}:
\begin{small}
\begin{equation} \label{def:recursive}
    \densityt{\pi}{P}{t}(s,a) = \int \densityt{\pi}{P}{t-1}(\tilde{s},\tilde{a}) P(s|\tilde{s},\tilde{a})  \pi(a|s) d\nu(\tilde{s},\tilde{a}),
\end{equation}
\end{small}
\hspace{-3pt}where $\nu$ is the Lebesgue measure. 

In the batch learning setting, we are given a dataset $D = \{(s_i, a_i, s'_i)\}_{i=1}^n$, where $s_i \sim d_{\pi_b}(s)$, $a_i \sim \pi_b$, and $s'_i \sim P(\cdot|s_i,a_i)$, where $\pi_b$ is some behavior policy that collects the data. For convenience, we write $(s,a,s') \sim D_{\pi_b} P$, where $D_{\pi_b}(s,a) = d_{\pi_b}(s) \pi_b(a|s)$. 
Let $E[\cdot]$ denote exact expectation and $E_n[\cdot]$ the empirical approximation using the $n$ data points of $D$.

Finally, we also need three classes $\W, \V, \M$ of functions. $\W \subset (\X \to \R)$ represents ratios between state-action occupancies, $\V \subset  (\State \to \R)$  represents value functions and $\M \subset (\X \to \Delta(\State))$ represents the class of models (or simulators) of the true environment. 

\textbf{Note}. Any Lemmas or Theorems presented without proof have full proofs in the Appendix.

\section{Minimax Model Learning (MML) for Off-Policy  Evaluation (OPE)}
\label{sec:mml}


\subsection{Natural Derivation}
We start with the off-policy evaluation (OPE) learning objective and derive the MML loss (Def \ref{loss:mml0}). In Section \ref{sec:learning}, we  show the loss also bounds off-policy optimization (OPO) error through its connection with OPE. 

\textbf{OPE Decision Problem.} 
The OPE objective is to estimate:
\begin{equation}\label{def:opeDP}
    J(\pi, P^\ast) \equiv E\left[\sum_{i=0}^\infty \gamma^i r_i \bigg| \substack{
s_0 \sim d_0 \\ 
a_i \sim \pi(\cdot|s_i) \\
s_{i+1} \sim P^\ast(\cdot|s_i,a_i)\\
r_i \sim \mathcal{R}(\cdot|s,a) \\
}
\right] ,
\end{equation} the performance of an evaluation policy $\pi$ in the true environment $P^\ast$, using only logging data $D$ with samples from $D_{\pi_b}P^\ast$. Solving this objective is difficult because the actions in our dataset were chosen with $\pi_b$ rather than $\pi$.  Thus, any $\pi \neq \pi_b$ potentially induces a ``shifted'' state-action distribution $D_{\pi} \neq D_{\pi_b}$, and ignoring this  shift can lead to poor estimation.

\textbf{Model-Based OPE.} 
Given a model class $\M$ and a desired evaluation policy $\pi$, we want to find a simulator $\widehat{P} \in \M$ using only logging data $D$ such that:
\begin{equation}\label{def:MBOPE}
    \widehat{P} = \arg\min_{P \in \M} |J(\pi, P) - J(\pi, P^\ast)|.
\end{equation} 
Interpreting Eq.~(\ref{def:MBOPE}), we run $\pi$ in $P$ to compute $J(\pi, P)$ as a proxy to $J(\pi, P^\ast)$.  If we find some $P \in \M$ such that $|\delta_{\pi}^{P,P^\ast}| = |J(\pi, P) - J(\pi, P^\ast)|$ is small, then $P$ is a good simulator for $P^\ast$.
%

\textbf{Derivation.} Using (\ref{def:JV}) and (\ref{def:JW}), we have:
\begin{align*}
\delta_{\pi}^{P,P^\ast} &= J(\pi, P) - J(\pi, P^\ast) \\
&= E_{s \sim d_0}[\val{\pi}{P}(s)] - E_{(s,a) \sim \density{\pi}{P^\ast} (\cdot,\cdot)}[ E_{ r \sim \mathcal{R}(\cdot|s,a)}[r]].
\end{align*}
Adding and subtracting $E_{(s,a) \sim \density{\pi}{P^\ast}}[\val{\pi}{P}(s)]$, we have:
\begin{align}
\delta_{\pi}^{P,P^\ast} = \ & E_{s \sim d_0}[\val{\pi}{P}(s)] -  E_{(s,a) \sim \density{\pi}{P^\ast}}[\val{\pi}{P}(s)] \label{eq:der_first} \\
&+ E_{(s,a) \sim \density{\pi}{P^\ast}}[\val{\pi}{P}(s) - E_{r\sim \mathcal{R}(\cdot|s,a)}[r]]. \label{eq:der_second}
\end{align}
To simplify the above expression, we make the following observations.
First, Eq.~\eqref{eq:der_second} can be simplified through the Bellman equation from Eq.~\eqref{def:Bellman}. To see this, notice that $\density{\pi}{P^\ast}$ is equivalent to some $d(s) \pi(a|s)$ for an appropriate choice of $d(s)$. Thus, 
\begin{align*}
E_{(s,a) \sim \density{\pi}{P^\ast}}&[\val{\pi}{P}(s) - E_{r\sim \mathcal{R}(\cdot|s,a)}[r]] \\
&= E_{s \sim d(\cdot)}[E_{a \sim \pi(\cdot|s)}[\val{\pi}{P}(s) - E_{r\sim \mathcal{R}(\cdot|s,a)}[r]]] \\
&= E_{s \sim d(\cdot)}[E_{a \sim \pi(\cdot|s)}[E_{s' \sim P(\cdot|s,a)}[\gamma \val{\pi}{P}(s)]]] \\
&= \gamma E_{(s,a) \sim \density{\pi}{P^\ast}}[E_{s' \sim P(\cdot|s,a)}[\val{\pi}{P}(s')]]. 
\end{align*}
Second, we can manipulate Eq.~\eqref{eq:der_first} using the definition 
\vspace{-.1cm} 
of $\density{\pi}{P}$ and  recursive property of $\densityt{\pi}{P}{t}$ from Eq. \eqref{def:recursive}:
\begin{small}
\begin{align*}
&E_{s\sim d_0}[\val{\pi}{P}(s)] - E_{(s,a) \sim \density{\pi}{P^\ast}}[\val{\pi}{P}(s)] \\
&=- \sum_{t=1}^\infty \gamma^t \int \densityt{\pi}{P^\ast}{t}(s,a) \val{\pi}{P}(s) d\nu(s,a) \\
&=- \gamma \sum_{t=0}^\infty \gamma^t \int \densityt{\pi}{P^\ast}{t+1}(s,a) \val{\pi}{P}(s) d\nu(s,a) \\
&=- \gamma \sum_{t=0}^\infty \gamma^t \int \densityt{\pi}{P^\ast}{t}(\tilde{s},\tilde{a}) P^\ast(s|\tilde{s},\tilde{a}) \pi(a|s) \val{\pi}{P}(s) d\nu(\tilde{s},\tilde{a},s,a) \\
&=- \gamma \sum_{t=0}^\infty \gamma^t \int \densityt{\pi}{P^\ast}{t}(s,a) P^\ast(s'|s,a) \val{\pi}{P}(s') d\nu(s,a,s') \\
&=- \gamma E_{(s,a) \sim \density{\pi}{P^\ast}}[ E_{s'\sim P^\ast(\cdot|s,a)}[\val{\pi}{P}(s')]].
\end{align*}
Combining the above allows us to succinctly express:
\begin{align*}
&\delta_{\pi}^{P,P^\ast} = \gamma E_{(s,a) \sim \density{\pi}{P^\ast}}[E_{s' \sim P(\cdot|s,a)}[\val{\pi}{P}(s')]]
\\ &\quad\quad\quad\quad - \gamma E_{(s,a) \sim \density{\pi}{P^\ast}}[ E_{s'\sim P^\ast(\cdot|s,a)}[\val{\pi}{P}(s')]].
\end{align*}
\end{small}
\hspace{-4pt}Since $D$ contains samples from $D_{\pi_b}$ and not $\density{\pi}{P^\ast}$, we use importance sampling to simplify the right-hand side of $\delta_{\pi}^{P,P^\ast}$ to: 
\begin{small}
\begin{equation}
\gamma   \substack{
\scalebox{1.1}{$E$} \\
(s,a,s') \sim D_{\pi_b} P^\ast \\
}
\left[ 
\frac{\density{\pi}{P^\ast}}{D_{\pi_b}} \left(
\substack{
\scalebox{1.1}{$E$} \\
\tilde{s} \sim P(\cdot|s,a) \\
}
[\val{\pi}{P}(\tilde{s})]
 - \val{\pi}{P}(s')\right)\right]. \label{eq:der_last}
\end{equation}
\end{small}


\begin{figure}[t]
  \includegraphics[width=\linewidth]{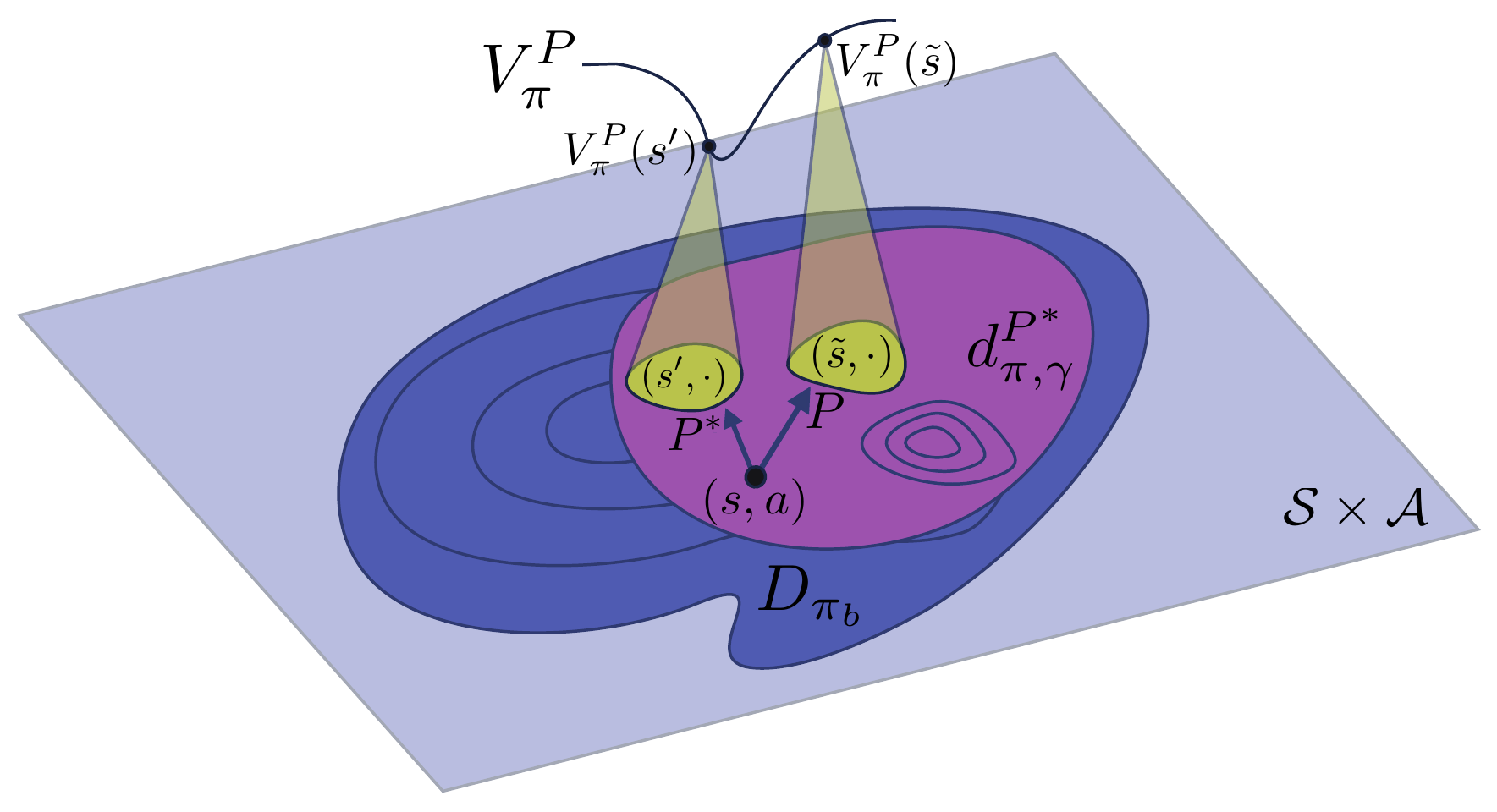}
\vspace{-0.3in}
\caption{\textit{Visual of Eq.~\eqref{eq:der_last}. The error at every point $(s,a)$ in $D_{\pi_b}$ is the difference between
$V^{P}_{\pi}(\tilde{s})$ (induced by following $P$) and $V^{P}_{\pi}(s')$ (induced by following $P^\ast$). We re-weight the points $(s,a)$ in $D_{\pi_b}$ to 
mimic
$d_{\pi,\gamma}^{P^\ast}$. Accumulating the errors exactly yields the OPE error of using $P$ as a simulator. MLE, instead, finds a $P$ ``pointing'' in the same direction as $P^\ast$  for all points in $D_{\pi_b}$, ignoring the discrepancy with $d_{\pi,\gamma}^{P^\ast}$. }}
 \label{fig:OPEObjective}
 \vspace{-.1in}
\end{figure}

%

\vspace{-.7cm}
Define $w_{\pi}^{P}(s,a) \equiv \frac{\density{\pi}{P}(s,a)}{D_{\pi_b}(s,a)}$. If we knew $w_{\pi}^{P^\ast}(s,a)$ and $\val{\pi}{P}$ (for every $P \in \M$), then we can select a $P \in \M$ to directly control 
$\delta_\pi^{P,P^\ast}$.  We encode this intuition as:



\begin{defn}\label{loss:mml0}[MML Loss] $\forall w \in \W, V \in \V, P \in \M$,
\begin{align*} 
    \Loss_{MML}(w,V,P) = &E_{(s,a,s') \sim D_{\pi_b}(\cdot,\cdot) P^\ast(\cdot|s,a)} [w(s,a) \;\cdot
    \\&\left( E_{\tilde{s} \sim P(\cdot|s,a)}[V(\tilde{s})] - V(s') \right)].
\end{align*}
When unambiguous, we will drop the MML subscript.
\end{defn}

Here we have replaced  $w_{\pi}^{P^\ast}(s,a)$
with $w$ coming from function class $\W$ and $\val{\pi}{P}$ with $V$ from class $\V$. The function class $\W$ represents the possible distribution shifts, while $\V$ represents the possible value functions.

%
With this intuition, we can formally guarantee that $J(\pi, P) \approx J(\pi, P^\ast)$ under the following \emph{realizability conditions}:

\begin{assumption}[Adequate Support]\label{assump:adequate support}
$D_{\pi_b}(s,a) > 0$ whenever $\density{\pi}{P}(s,a) > 0$. Define $w^P_\pi(s,a) \equiv \frac{\density{\pi}{P}(s,a)}{D_{\pi_b}(s,a)}$. 
\end{assumption}

\begin{assumption}[OPE Realizability]\label{assump:ope_realizability}
For a given $\pi$, $\W \times \V$ contains at least one of $(w^P_\pi, \val{\pi}{P^{\ast}})$ or $(w^{P^\ast}_\pi, \val{\pi}{P})$ for every $P \in \M$.
\end{assumption}


\begin{thm}[MML \& OPE] \label{thm:opeerror}
Under Assumption \ref{assump:ope_realizability},
\begin{small}
\begin{equation}
    |J(\pi,\widehat{P}) - J(\pi,P^\ast)| \leq \gamma \min_{P \in \M} \max_{w \in \W, V \in \V} |\Loss(w,V,P)|,
\end{equation}
\end{small}
\hspace{-4pt}where $\widehat{P} = \arg\min_{P \in \M} \max_{w\in\W, V\in\V} |\Loss(w,V,P)|$.
\end{thm}
\begin{rem}
We want to choose $\V, \W, \M$ carefully so that many $P \in \M$ satisfy $\Loss(w,V,P) = 0$ and Assumption \ref{assump:ope_realizability}. 
By inspection, $\Loss(w,V,P^\ast) = 0$ for any  $V \in \V, w \in \W$.
\end{rem}
\begin{rem} \label{rem:verifiability}
While $V^{P}_{\pi} \in \V \; \forall P \in \M$ appears strong, it can be verified for every $P\in \M$ before accessing the data, as the condition does not depend on $P^\ast$. In principle, we may redesign $\V$ to guarantee this condition.
\end{rem}
\begin{rem} When $\gamma = 0$,  $J$ does not depend on a transition function, so $J(\pi,P) = J(\pi,P^\ast) \;  \forall P \in \M$.
\end{rem}

$\Loss(w,V,P^\ast)=0$ and Theorem \ref{thm:opeerror} implies that the following learning procedure will be robust to any distribution shift in $\W$ and any value function in $\V$:
\begin{defn}[Minimax Model Learning (MML)]
\begin{equation}
    \widehat{P} = \arg\min_{P \in \M} \max_{w\in\W, V\in\V} |\Loss_{MML}(w,V,P)|. \label{phat}
\end{equation}
\end{defn}

\subsection{Interpretation and Verifiability} 

Figure \ref{fig:OPEObjective} gives a visual illustration of Eq. \eqref{eq:der_last} which leads to the MML Loss (Def \ref{loss:mml0}). 
$\pi_b$ has induced an ``inbalanced'' training dataset $D_{\pi_b}$ and the importance sampling term acts to rebalance our data because our test dataset will be $\density{\pi}{P^\ast}$, induced by $\pi$. Because the objective is OPE, we don't mind that $\hat{P}$ is different than $P^\ast$ so long as $E_{\hat{P}}[V_{\pi}^{\hat{P}}] \approx E_{P^\ast}[V_{\pi}^{\hat{P}}]$. In other words, the size of $V_{\pi}^{\hat{P}}$ tells us which state transitions are important to model correctly. We want to appropriately utilize the capacity of our model class $\M$ so that $\hat{P}$ models $P^\ast$ when $V_{\pi}^{\hat{P}}$ is large. When it is small, it may be better off to ignore the error in favor of other states.

Theorem \ref{thm:opeerror} quantifies the error incurred by evaluating $\pi$ in $\widehat{P}$ instead of $P^\ast$, assuming Assumption \ref{assump:ope_realizability} holds. For OPE, $\widehat{P}$ is a reasonable proxy for $P$. In this sense, MML is a principled method approach for model-based OPE.  See Appendix \ref{app:ope:sec:main} for a complete proof of Thm \ref{thm:opeerror} and Appendix \ref{app:ope:sec:samplecomplexity} for the sample complexity analysis.


If the exploratory state distribution $d_{\pi_b}$ and $\pi_b$ are known then $D_{\pi_b}$ is known. 
In this case, we can also verify that $w_{\pi}^{P} \in \W$ for every $P \in \M$ a priori.  Together with Remark \ref{rem:verifiability}, we may assume that both  $w_{\pi}^{P} \in \W$ and $V^{P}_{\pi} \in \V$ for all $P \in \M$.  Consequently, only one of $V^{P^\ast}_{\pi} \in \V$ or $w_{\pi}^{P^\ast} \in \W$ has to be realizable for Theorem \ref{thm:opeerror} to hold. 

Instead of checking for realizability apriori, we can perform post-verification that $w_{\pi}^{\widehat{P}} \in \W$ and $V^{\widehat{P}}_{\pi} \in \V$.  Together with the terms depending on $P^\ast$, realizability of these are also sufficient for Theorem \ref{thm:opeerror} to hold.  This relaxes the strong ``for all $P \in \M$'' condition.


\subsection{Comparison to Model-Free OPE}
\label{sec:MFOPEComparison}

Recent model-free OPE literature \citep[e.g.,][]{liu2018breaking, uehara2019minimax} has similar realizability  assumptions to Assumption \ref{assump:ope_realizability}.  

As an example, the method MWL \citep{uehara2019minimax} takes the form of:
\begin{align*}
    &J(\pi, P^\ast) \approx  E_{(s,a,r) \sim D_{\pi_b} }[\widehat{w}(s,a) r] \\
    \text{where} \;\; &\widehat{w} = \arg\min_{w \in \W} \max_{Q \in \Q} | \Loss_{MWL}(w,Q)|, 
\end{align*}
requiring $Q_{\pi}^{P^\ast}$ to be realized to be a valid upper bound.
Here $\Q$ is analogous to our function class $\V$ where $E_{a \sim \pi(a|s)}[Q_{\pi}^{P^\ast}(s,a)] = \val{\pi}{P^\ast}(s).$ The loss $\Loss_{MWL}$ has no dependence on $P$ and is therefore model-free. MQL \citep{uehara2019minimax}  has analogous realizability conditions to MWL. 

Our loss, $\Loss_{MML}$, has the same realizability assumptions in addition to one related to $\M$ (and not $\M^\ast$). As discussed in Remark~\ref{rem:verifiability}, these $\M$-related assumptions can be verified a priori and in principle, satisfied by redesigning the function classes. Therefore, they do not pose a substantial theoretical challenge. See Section \ref{sec:experiments} for a practical discussion.

An advantage of model-free approaches is that when both $w^{P^\ast}_{\pi}, Q_{\pi}^{P^\ast}$ are realized,  they return an exact OPE point estimate.  In contrast, MML additionally requires some $P \in \M$ that makes the loss zero for any $w \in \W, V \in \V$. The advantage of MML is the increased flexibility of a model, enabling OPO (Section \ref{sec:learning}) and visualization of results through simulation (leading to more transparency).

While recent model-free OPE and our method both take a minimax approach, the classes $\W, \V, \M$ play different roles. In the model-free case, minimization is w.r.t either $\W$ or $\V$ and maximization is w.r.t the other. In our case, $\W,\V$ are on the same (maximization) team, while minimization is over $\M$. This allows us to treat $\W \times \V$ as a single unit, and represents distribution-shifted value functions. A member of this class, $E_{\text{data}}[wV]$ ($= E_{(s,a) \sim D_{\pi_b}}[\frac{\density{\pi}{P^\ast}}{D_{\pi_b}} \val{\pi}{P}(s)]$),  ties together the OPE estimate.  

\subsection{Misspecification of $\M,\V,\W$}
\label{sec:misspecification}



Suppose Assumption \ref{assump:ope_realizability} does not hold and $P^\ast \not\in \M$. Define a new function $h(s,a,s') \in \Hh = \{w(s,a)V(s') | (w,V) \in \W \times \V \}$ then we redefine $\Loss$:
\begin{align*}
    \Loss(h,P) = &E_{(s,a,s') \sim D_{\pi_b}(\cdot,\cdot) P^\ast(\cdot|s,a)} [ \\
    &E_{x\sim P(\cdot|s,a)}[h(s,a,x)] - h(s,a,s')].
\end{align*}


\begin{prop}[Misspecification discrepancy for OPE]\label{lem:misspec} Let $\Hh \subset (\State \times \A \times \State \to \R)$ be a set of functions on $(s,a,s')$. Denote $(WV)^\ast = w_{\pi}^{P^\ast}(s,a)V^{P}_{\pi}(s')$ (or, equivalently, $(WV)^\ast = w_{\pi}^{P}(s,a)V^{P^\ast}_{\pi}(s')$). 
\begin{equation}
    |J(\pi,\widehat{P}) - J(\pi,P^\ast)| \leq \gamma \min_P \max_{h \in \Hh}  |\Loss(h,P)| + \gamma \epsilon_\Hh,
\end{equation}
where $\epsilon_\Hh = \max_{P \in \M} \min_{h \in \mathcal{H}} |\Loss((WV)^\ast - h, P)|.$

\end{prop}

$\Loss(WV^\ast - h, P)$ measures the difference between $h$ and $(WV)^\ast$. Another interpretation of Prop \ref{lem:misspec} is if $\arg\max_{\Hh \cup \{(WV)^\ast\}} \Loss(h,P) = (WV)^\ast$ for some $P \in \M$ then MML returns a value $\gamma \epsilon_\Hh$ below the true upper bound, otherwise the output of MML remains the upperbound. This result illustrates that realizability is sufficient but not necessary for MML to be an upper-bound on the loss.

\subsection{Application to the Online Setting}

While the main focus of MML is batch OPE and OPO, we will make a few remarks relating to the online setting. In particular, if we assume we can engage in online data collection then $\W = \{ \mathbf{1} \}$ (the constant function), representing no distribution shift since $\pi_b = \pi$. When VAML and MML share the same function class $\V$, we can show that $\min_\M \max_{\W, \V} \Loss_{MML}(w, V, P)^2 \leq \min_P \Loss_{VAML}(\V, P)$ for any $\V, \M$. In other words, MML is a tighter decision-aware loss even in online data collection. In addition, MML enables greater flexibility in the choice of $\V$. See Appendix \ref{app:ope:sec:batch2online} for further details.

\section{Off-Policy Optimization (OPO)}
\label{sec:learning}

\subsection{Natural Derivation}

In this section we examine how our MML approach can be integrated into the policy learning/optimization objective. In this setting, the goal is to find a good policy with respect to the true environment $P^\ast$ without interacting with $P^\ast$. 

\textbf{OPO Decision Problem.} 
Given a policy class $\Pi$ and access to only a logging dataset $D$ with samples from $D_{\pi_b}P^\ast$, find a policy $\pi \in \Pi$ that is competitive with the unknown optimal policy $\pi^\ast_{P^\ast}$:
\begin{equation}\label{def:opoDP}
    \widehat{\pi}^\ast = \arg\min_{\pi \in \Pi} |J(\pi, P^\ast) - J(\pi^\ast_{P^\ast}, P^\ast)|.
\end{equation} 
\textbf{Note:} No additional exploration is allowed.

\textbf{Model-Based OPO.} 
Given a model class $\M$, we want to find a simulator $\widehat{P} \in \M$ using only logging data $D$ and subsequently learn $\pi^\ast_{\widehat{P}} \in \Pi$ in $\widehat{P}$ through any policy optimization algorithm which we call Planner($\cdot$). 
 \begin{algorithm}[H]
	\caption{Standard Model-Based OPO } 
	\label{algo:standardopo}
	\begin{algorithmic}[1]
    	\REQUIRE $D = D_{\pi_b}P^\ast$, Modeler, Planner
	    \STATE Learn $\widehat{P} \leftarrow$ Modeler($D$)
	    \STATE Learn $\widehat{\pi}^\ast_{P} \leftarrow \text{Planner}(\widehat{P})$
	    \RETURN $\widehat{\pi}^\ast_{P}$
	\end{algorithmic}
\end{algorithm}
In Algorithm \ref{algo:standardopo}, Modeler($\cdot$) refers to any (batch) model learning procedure. The hope for model-based OPO is that the ideal in-simulator policy $\pi^\ast_{\widehat{P}}$ and the actual best (true environment) policy $\pi^\ast_{P^\ast}$ perform competitively: $J(\pi^\ast_{\widehat{P}}, P^\ast) \approx J(\pi^\ast_{P^\ast}, P^\ast)$. Hence, instead of minimizing Eq \eqref{def:opoDP} over all $\pi \in \Pi$, we can focus $\Pi = \{\pi^\ast_{P}\}_{P \in \M}$.


\textbf{Derivation.} Beginning with the objective, we add zero twice:
\begin{align*}
    J(&\pi^\ast_{P^\ast}, P^\ast) - J(\pi^\ast_{P}, P^\ast) =
    \underbrace{J(\pi^\ast_{P^\ast}, P^\ast) - J(\pi^\ast_{P^\ast}, P)}_{(a)}  \\
    &\quad+  \underbrace{J(\pi^\ast_{P^\ast}, P) - J(\pi^\ast_{P}, P)}_{(b)}  + \underbrace{J(\pi^\ast_{P}, P) - J(\pi^\ast_{P}, P^\ast)}_{(c)}.
\end{align*}
Term (b) is non-positive since $\pi^\ast_P$ is optimal in $P$ ($\pi^\ast_{P^\ast}$ is suboptimal), so we can drop it in an upper bound. 
Term (a) is the OPE estimate of $\pi^\ast_{P^\ast}$ and term (c) the OPE estimate of $\pi^\ast_{P}$, implying that we should use Theorem \ref{thm:opeerror}. With this intuition, we have:

\begin{thm}[MML \& OPO]\label{thm:learningexact}
If $w^{P^\ast}_{\pi^\ast_{P^\ast}},w^{P^\ast}_{\pi^\ast_{P}} \in \mathcal{W}$ and $V^{P}_{\pi^\ast_{P^\ast}}, V^{P}_{\pi^\ast_{P}} \in \mathcal{V}$ for every $P \in \M$ then:
\begin{equation*}
    |J(\pi^\ast_{P^\ast}, P^\ast) - J(\pi^\ast_{\widehat{P}}, P^\ast)|  \leq 2 \gamma \min_P \max_{w,V} | \Loss(w,V,P) |.
\end{equation*} 
The statement also holds if, instead, $w^{P}_{\pi^\ast_{P^\ast}},w^{P}_{\pi^\ast_{P}} \in \mathcal{W}$ and $V^{P^\ast}_{\pi^\ast_{P^\ast}}, V^{P^\ast}_{\pi^\ast_{P}} \in \mathcal{V}$ for every $P \in \M$.
\end{thm}


\subsection{Interpretation and Verifiability}

Theorem \ref{thm:learningexact} compares two different policies in the same (true) environment, since $\pi^\ast_{\widehat{P}}$ will be run in $P^\ast$ rather than $\widehat{P}$. In contrast, Theorem \ref{thm:opeerror} compared the same policy in two different environments. 
The derivation of Theorem \ref{thm:learningexact} (see Appendix \ref{app:opo:sec:main}) shows that having a good bound on the OPE objective is sufficient for OPO. MML shows how to learn a model that exploits this relationship.


Furthermore, the realizability assumptions of Theorem \ref{thm:learningexact} relax the requirements of an OPE oracle. Rather than requiring the OPE estimate for every $\pi$, it is sufficient to have the OPE estimate of $\pi_{P^\ast}^\ast$ and $\pi_{P}^\ast$ (for every $P \in \M$) when there is a $P \in \M$ such that $\Loss(w,V,P)$ is small for any $w \in \W, V \in \V$. 

We could have instead examined the quantity $\min_\pi |J(\pi^\ast_{P^\ast},P^\ast) - J(\pi, P^\ast)|$ directly from Eq \eqref{def:opoDP}. What we would find is that the upper bound is $2 \min_P \max_{w,V} |E_{d_0}[V] - \Loss(w,V,P)|$ and the realizability requirements would be that $V^P_{\pi} \in \V, w^{P^\ast}_{\pi} \in \W$ for every $\pi$ in some policy class. This is a much stronger requirement than in Theorem \ref{thm:learningexact}. 


For OPO, apriori verification of realizability is possible by enumerating over $P \in \M$.
Whereas the target policy $\pi$ was fixed in OPE, now $\pi^\ast_{P}$ varies for each $P \in \M$. It may be more practical to, as in OPE, perform post-verification that  $w^{P}_{\pi^\ast_{\widehat{P}}} \in \mathcal{W}$ and $V^{P}_{\pi^\ast_{\widehat{P}}} \in \mathcal{V}$. If they do not hold, then we can modify the function classes until they do.   This relaxes the ``for every $P \in \M$" condition and  leaves only a few unverifiable quantities relating to $P^\ast$.


Sample complexity and function class misspecification results for OPO can be found in Appendix  \ref{app:opo:sec:samplecomplexity}, \ref{app:opo:sec:misspec}.

\subsection{Comparison to Model-Free OPO}


For minimax model-free OPO, \cite{chen2019FQI} have analyzed a minimax variant of Fitted Q Iteration (FQI) \citep[]{Ernst2005FQI}, inspired by \citet{antos2008learning}. FQI is a commonly used model-free OPO method. In addition to realizability assumptions, these methods also maintain a completeness assumption: the function class of interest is closed under bellman update. Increasing the function class size can only help realizability but may break completeness. It is unknown if the completeness assumption of FQI is removable  \citep[]{chen2019FQI}. MML only has realizability requirements.



\makeatletter
\newcommand{\customlabel}[3]{%
   \protected@write \@auxout {}{\string \newlabel {#1}{{\ref{#2}~(#3)}{\thepage}{}{}{}} }%
}
\makeatother

\section{Scenarios \& Considerations}
\label{caseStudies}

In this section we investigate a few scenarios where we can calculate the class $\V$ and $\W$ or modify the loss based on structural knowledge of $\M, \W,$ and $\V$. 

In examining the scenarios, we aim to verify that MML gives \textit{sensible} results. For example, in scenarios where we know MLE to be optimal, MML should ideally coincide. Indeed, we show this to be the case for the tabular setting and Linear-Quadratic Regulators.  Other scenarios include showing that MML is compatible with incorporating prior knowledge using either a nominal dynamics model or a kernel.

The proofs for any Lemmas in this section can be found in Appendix \ref{app:sec:scenarios}.  



\subsection{Linear \& Tabular Function Classes}

When $\W, \V, \M$ are linear function classes then the entire minimax optimization has a closed form solution. In particular, $\M$ takes the form $P = \phi(s,a,s')^T \alpha$ where $\phi \in \mathbb{R}^{|\State \times \A \times \State|}$ is some basis of features with $\alpha \in \mathbb{R}^{|\State \times \A \times \State|}$ its parameters and $(w(s,a), V(s')) \in \W\V = \{\psi(s,a,s')^T\beta : \|\beta\|_\infty < +\infty\}$ where $\psi \in \mathbb{R}^{|\State \times \A \times \State|}$. 

\begin{prop}[Linear Function classes] \label{lem:linearfunc}
Let $P = \phi(s,a,s')^T \alpha $ where $\phi \in \R^{|\State \times \A \times \State|}$ is some basis of features with $\alpha$ its parameters. Let $(w(s,a), V(s')) \in \W\V = \{\psi(s,a,s')^T\beta : \|\beta\|_\infty < +\infty\}$. Then, 
\begin{equation}
    \widehat{\alpha} = E_n^{-T}\left[\int \phi(s,a,s') \psi(s,a,s')^T d\nu(s') \right] E_n[\psi(s,a,s')],
\end{equation}
if $E_n\left[\int \phi(s,a,s') \psi(s,a,s')^T d\nu(s') \right]$ has full rank.
\end{prop}

The tabular setting, when the state-action space is finite, is a common special case. We can choose: \begin{equation}\label{tabular_repr}
    \psi(s,a,s') = \phi(s,a,s') = e_{i}
\end{equation} as the $i$th standard basis vector where $i = s|\A||\State| + a|\State| + s'$.  There is no model misspecification in the tabular setting (i.e., $P^\ast \in \M$), therefore $\widehat{P} = P^\ast$ in the case of infinite data.

\begin{prop}[Tabular representation] \label{lem:tabular}
Let $P = \phi(s,a,s')^T \alpha $ with $\phi \in \R^{|\State \times \A \times \State|}$ as in Eq \eqref{tabular_repr} and $\alpha$ its parameters. Let $(w(s,a), V(s')) \in \W\V = \{\phi(s,a,s')^T\beta : \|\beta\|_\infty < +\infty \}$. Assume we have at least one data point from every $(s,a)$ pair. Then:
\begin{equation}
    \widehat{P}_n(s'|s,a)= \frac{\#\{(s,a,s') \in D\}}{\#\{(s,a,\cdot) \in D \}}.
\end{equation}
\end{prop}

Prop. \ref{lem:tabular} shows that MML and MLE coincide, even in the finite-data regime. Both models are simply the observed propensity of entering state $s'$ from tuple $(s,a)$. 


\subsection{Linear Quadratic Regulator (LQR)}\label{subsec:lqr}

The Linear Quadratic Regulator (LQR) is defined as linear transition dynamics $P^\ast(s' | s,a) = A^\ast s + B^\ast a + w^\ast$ where $w^\ast$ is random noise and a quadratic reward function $\mathcal{R}(s,a) = s^T Q s + a^T R a$ for $Q, R \geq 0$ symmetric positive semi-definite. For ease of exposition we assume that $w^\ast \sim N(0, \sigma^{\ast 2} I)$.  We assume that $(A^\ast, B^\ast)$ is controllable. Exploiting the structure of this problem, we can check that every $V \in \V$ takes the form $V(s) = s^T U s + q$ for some symmetric semi-positive definite $U$ and constant $q$ (Appendix Lemma \ref{lem:VQuad_LQR}). 

Furthermore, we know controllers of the form $\pi(a|s) = - K s$ where $K \in \R^{k \times n}$ are optimal in LQR \citep[]{bertsekas2005dynamic}. We consider determistic and therefore \textit{misspecified} models of the form $P(s'|s,a) = As + Ba$.  $\W$ is a Gaussian mixture and we can write $\Loss_{MML}$ as a function of $U,K$ and $(A,B)$ (Appendix Lemma \ref{lem:L_LQR}).

\begin{prop}[MML + MLE Coincide for LQR]\label{lem:MML_MLE_LQR} Let $A \in \R^{n\times n}, B \in \R^{n \times k}, K \in \R^{k \times n}$. Let $U \in \mathcal{S}^{n}$ be positive semi-definite.  Set $k=1$, a single input system.  Then,
\begin{align*} 
\arg\min_{(A,B)}& \max_{K,U} |\Loss_{MML}(K,U,(A,B))| = (A^\ast, B^\ast)\\
&= \arg\min_{(A,B)} \Loss_{MLE}(A,B) .
\end{align*}
\end{prop}
Despite model misspecification, both MLE and MML give the correct parameters $(\widehat{A}, \widehat{B}) = (A^\ast, B^\ast)$. 
We leave showing that MML and MLE coincide in multi-input ($k > 1$) LQR systems for future work.

\subsection{Residual Dynamics \& Environment Shift}

Suppose we already had some baseline model $P_0$ of $P^\ast$. Alternatively, we may view this as the real world starting with (approximately) known dynamics $P_0$ and drifting to $P^\ast$. We can modify MML to incorporate knowledge of $P_0$ to find the residual dynamics:
\begin{defn}\label{loss:mml}[Residual MML Loss] For $w \in \W, V \in \V, P \in \M$,
\begin{align*} 
    &\Loss(w,V,P) = E_{(s,a,s') \sim D_{\pi_b}(\cdot,\cdot) P^\ast(\cdot|s,a)} [w(s,a) \;\cdot
    \\&\left( E_{x\sim P_0(\cdot|s,a)}[\frac{P_0(x|s,a)-P(x|s,a)}{P_0(x|s,a)} V(x)] - V(s') \right)].
\end{align*}
\end{defn}
This solution form matches the intuition that having prior knowledge in the form of $P_0$ focuses the learning objective on the difference between $P^\ast$ and $P_0$.

\begin{figure*}[ht!]
    \centering     
\hfill%
    \customlabel{fig:LQR_MML_vs_rest}{fig:lqr}{left}%
        \includegraphics[width=.45\linewidth]{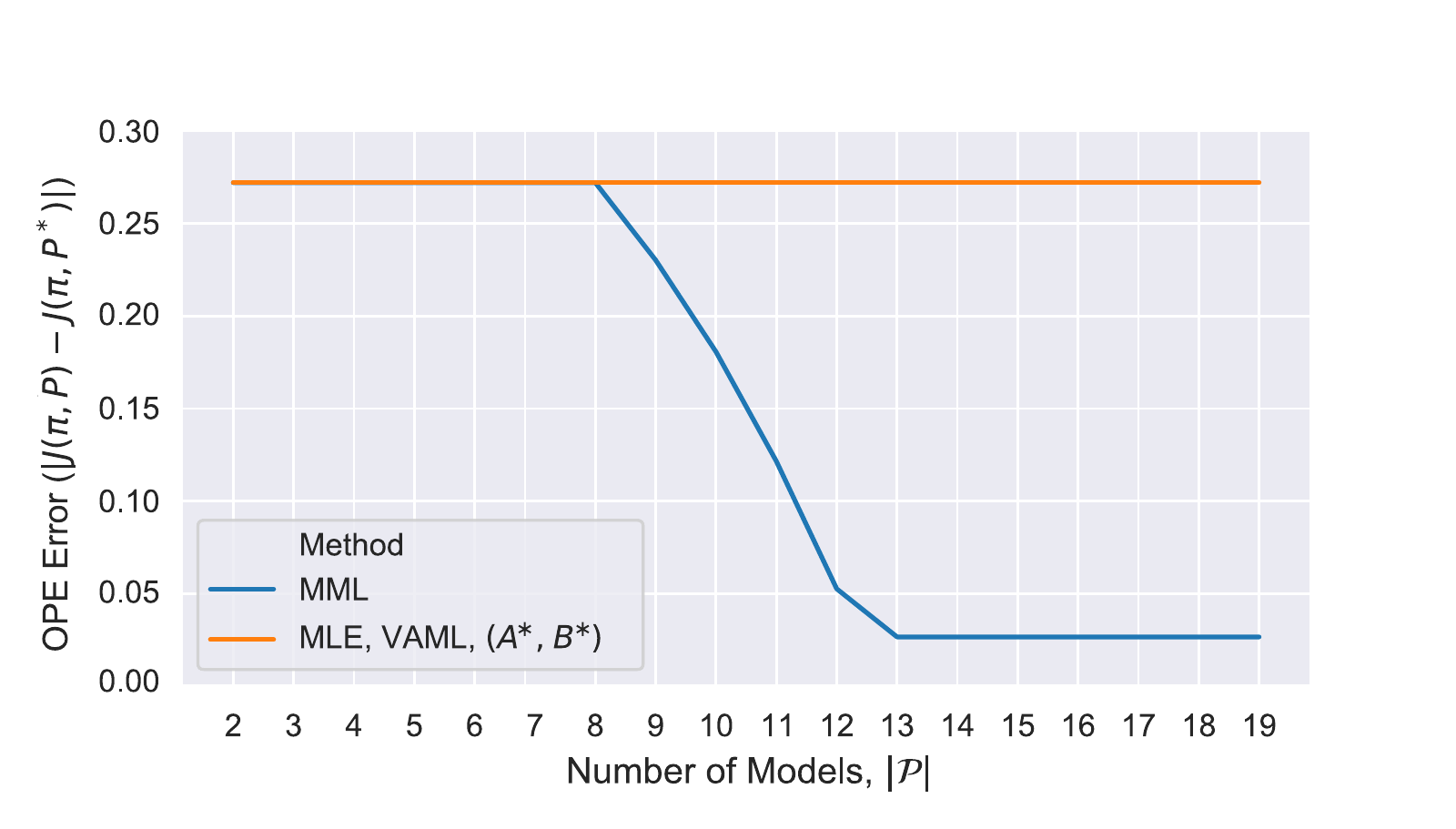}
\hfill%
    \customlabel{fig:LQR_verifiability}{fig:lqr}{right}%
        \includegraphics[width=.45\linewidth]{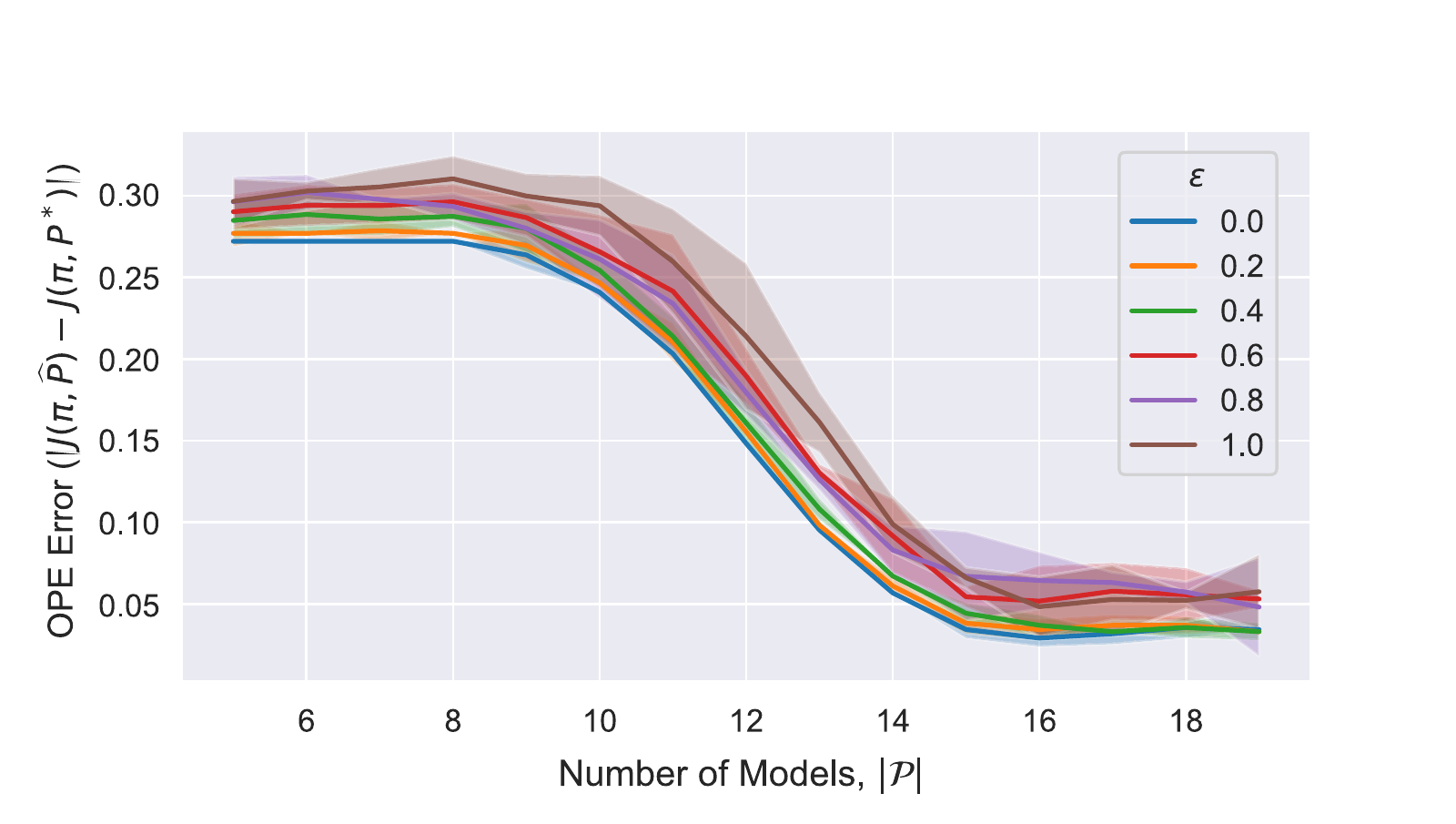}
        \vspace{-0.1in}
\caption{\emph{LQR}.
             \emph{(Left, OPE Error)} MML finds the $P \in \M$ with the lowest OPE error as $\M$ gets richer. Since calculations are done in expectation, no error bars are included. 
             \emph{(Right, Verifiability)} The OPE error (smoothed) increases with misspecification in $\V$ parametrized by $\epsilon$, the expected MSE between the true $V^{P^\ast}_{\pi} \not\in \V$ and the approximated $\widehat{V}^{P^\ast}_{\pi} \in \V$. Nevertheless, directionally they all follow the same trajectory as $\M$ gets richer.
             }
\label{fig:lqr}
\end{figure*}


\subsection{Incorporating Kernels}

Our approach is also compatible with incorporating kernels (which is a way of encoding domain knowledge such as smoothness) to learn in a Reproducing Kernel Hilbert Space (RKHS). For example, we may derive a closed form for $\max_{(w,V) \in \mathcal{WV}} \Loss(w,V,P)^2$ when $\W \times \V$ corresponds to an RKHS and use standard gradient descent to find $\widehat{P} \in \M$, making the minimax problem much more tractable. See Appendix \ref{app:scenarios:rkhs} for a detailed discussion on RKHS, computational issues relating to sampling from $P$ and alternative approaches to solving the minimax problem.

\makeatletter


\section{Experiments}
\label{sec:experiments}

In our experiments, we seek to answer the following questions: (1) Does MML prefer models that minimize the OPE objective? (2) What can we expect when we have misspecification in $\V$? (3) How does MML perform against MLE and VAML in OPE? (4) Does our approach complement modern offline RL approaches?  For this last question, we consider integrating MML with the recently proposed MOREL \citep[]{kidambi2020morel} approach for offline RL. See Appendix \ref{app:morel} for details on MOREL.

\subsection{Brief Environment Description/Setup}


We perform our experiments in three different domains.

\textbf{Linear-Quadratic Regulator (LQR).} The LQR domain is a 1D environment with stochastic dynamics $P^\ast(s'|s,a)$. We use a finite class $\M$ consisting of deterministic policies.
We ensure $V^P_{\pi} \in \V$ for all $P \in \M$ by solving the equations in Appendix Lemma \ref{lem:VQuad_LQR}. We ensure $W^{P^\ast}_{\pi} \in \W$ using Appendix Equation \eqref{eq:lqr_dpi}.

\textbf{Cartpole \citep[]{OpenAIgym}.} The reward function is modified to be a function of angle and location rather than 0/1 to make the OPE problem more challenging. Each $P \in \M$ is a parametrized NN that outputs a mean, and logvariance representing a normal distribution around the next state.
We model the class $\W\V$ as a RKHS as in Prop \ref{lem:rkhs} with an RBF kernel.

\textbf{Inverted Pendulum (IP) \citep[]{laypy}.}  This IP environment has a Runge-Kutta(4) integrator rather than Forward Euler (Runge-Kutta(1)) as in OpenAI \citep[]{OpenAIgym}, producing significantly more realistic data. Each $P \in \M$ is a deterministic model parametrized with a neural network. We model the class $\W\V$ as a RKHS as in Prop \ref{lem:rkhs} with an RBF kernel.

\textbf{Further Detail} A thorough description of the environments, experimental details, setup and hyperparameters can be found in Appendix \ref{app:experiments}.

\subsection{Results}

\textbf{Does MML prefer models that minimize the OPE objective?}  We vary the size of the model class Figure \ref{fig:LQR_MML_vs_rest} testing to see if MML will pick up on the models which have better OPE performance. When the sizes of $|\M|$ are small, each method selects $(A^\ast, B^\ast)$ (e.g. $P(s'|s,a) = A^\ast s + B^\ast a$), the deterministic version of the optimal model. However, as we increase the richness of $\M$, MML begins to pick up on models that are able to better evaluate $\pi$.

Two remarks are in order. In LQR, policy optimization in $(A^\ast, B^\ast)$ coincides with policy optimization in $P^\ast$. Therefore, if we tried to do policy optimization in our selected model then our policy would be suboptimal in $P^\ast$. Secondly, MML deliberately selects a model other than $(A^\ast, B^\ast)$ because a good OPE estimate relies on appoximating the contribution from the stochastic part of $P^\ast$.


There is a trade-off between the OPE objective and the OPO objective. MML's preference is dependent on the capacities of $\M, \W, \V$. Figure \ref{fig:LQR_MML_vs_rest} illustrates OPE
is preferred for $\W$ fixed. Appendix Figure \ref{fig:LQR_MML_vs_rest_OPO} explores the OPO objective and shows that if we increase $\W$ then OPO becomes favored. In some sense we are asking MML to be robust to many more OPE problems as $|\W|\uparrow$ and so the performance on any single one decreases, favoring OPO.

\textbf{What can we expect when we have misspecification in $\V$?} To check verifiability in practice, we would run $\pi$ in a few $P \in \M$ and calculate $V^{P}_\pi$. We would check if $V^{P}_\pi \in \V$ by fitting $\widehat{V}^P_{\pi}$ and measuring the empirical gap $E[(\widehat{V}_\pi^{P} - V_\pi^{P})^2] = \epsilon^2$.

Figure \ref{fig:LQR_verifiability} shows how MML performs when $V_\pi^{P} \not\in \V$ but we do have $\widehat{V}_\pi^{P}(s) = V_\pi^{P}(s) + \mathcal{N}(0,\epsilon) \in \V$. Since $E[(\widehat{V}_\pi^{P} - V_\pi^{P})^2] = \epsilon^2$ then $\epsilon$ is the root-mean squared error between the two functions. Directionally all of the errors go down as $|\M|\uparrow,$ however it is clear that $\epsilon$ has a noticeable effect. We speculate that if this error not distributed around zero and instead is dependent on the state then the effects can be worse.

\begin{figure}[!t]
  \includegraphics[width=\linewidth]{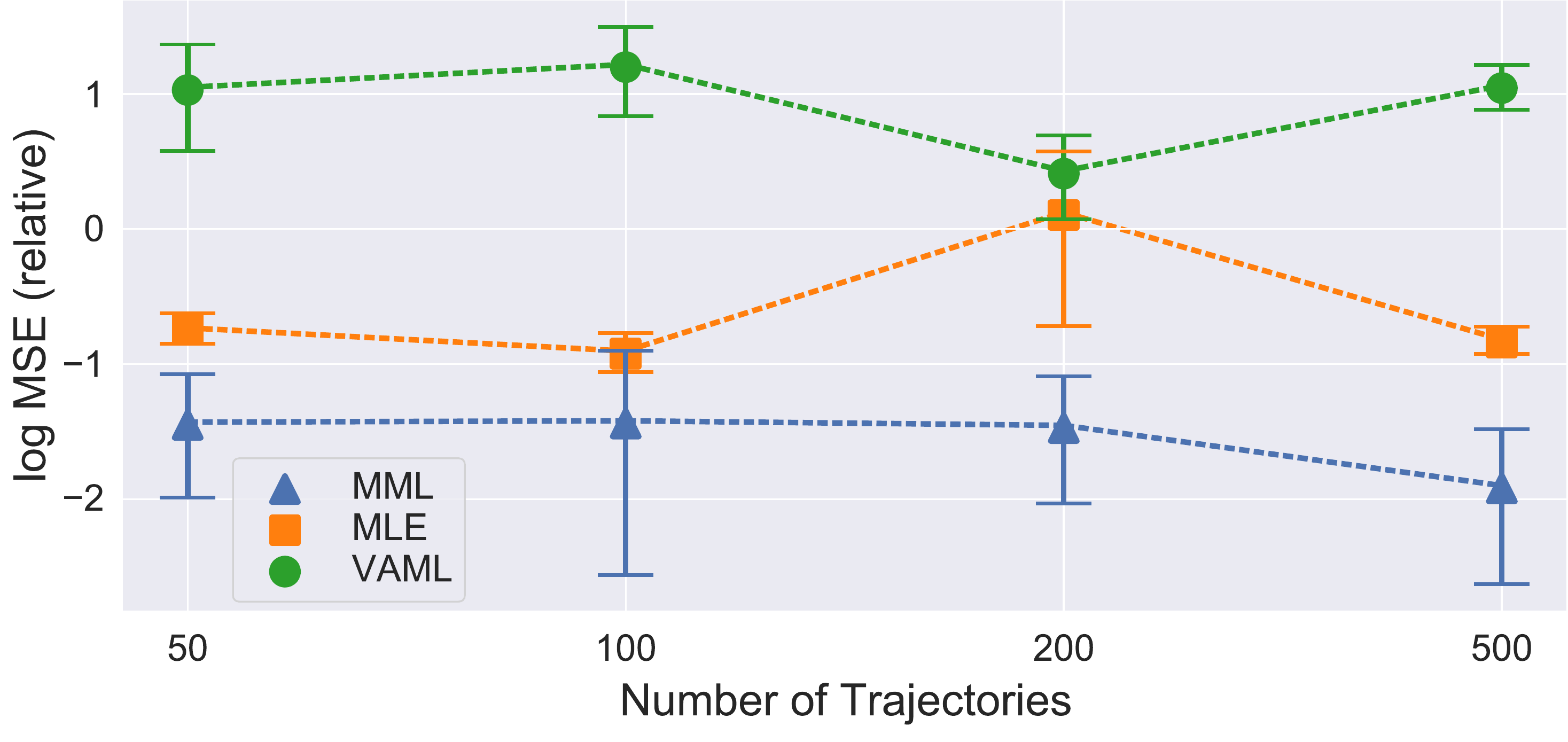}
\vspace{-0.3in}
\caption{\textit{ (Cartpole, OPE Error) Comparison of model-based approaches for OPE with function-approx.} \textit{Lower is better. MML outperforms others.
Not pictured: traditional model-free methods such as IS/PDIS have error of order 3-8.
}}
 \label{fig:OPE}
 \vspace{-.2in}
\end{figure}

\textbf{How does MML perform against MLE and VAML in OPE?} In addition to Figure \ref{fig:LQR_MML_vs_rest},  Figure \ref{fig:OPE} also illustrates that our method outperforms the other model-learning approaches in OPE. The environment and reward function is challenging, requiring function approximation.
Despite the added complexity of solving a minimax problem, doing so gives nearly an order of magnitude improvement over MLE and many orders over VAML. This validates that MML is a good choice for model-learning for OPE.


\vspace{-.05in}
 \begin{algorithm}[H]
	\caption{OPO Algorithm (based on MOREL \citep[]{kidambi2020morel}) }
	\label{algo:main_morel}
	\begin{algorithmic}[1]
    	\REQUIRE $D$, $\Loss$ among \{MML, MLE, VAML\}
	    \STATE Learn an ensemble of dynamics $P_1,\ldots,P_4 \in \M$ using $P_i = \arg\min_{P \in \M} \Loss(D)$
	    \STATE Construct a pessimistic MDP $\mathcal{M}$ (see Appendix \ref{app:morel}) with $P(s,a) = \frac{1}{4} \sum_{i=1}^{4} P_i(s,a)$.
	    \STATE $\widehat{\pi} \leftarrow \text{PPO}(\mathcal{M})$ (Best of 3) \citep[]{Schulman17PPO}
	\end{algorithmic}
\end{algorithm}

\begin{figure}[ht!]
\vspace{-0.27in}
  \includegraphics[width=\linewidth]{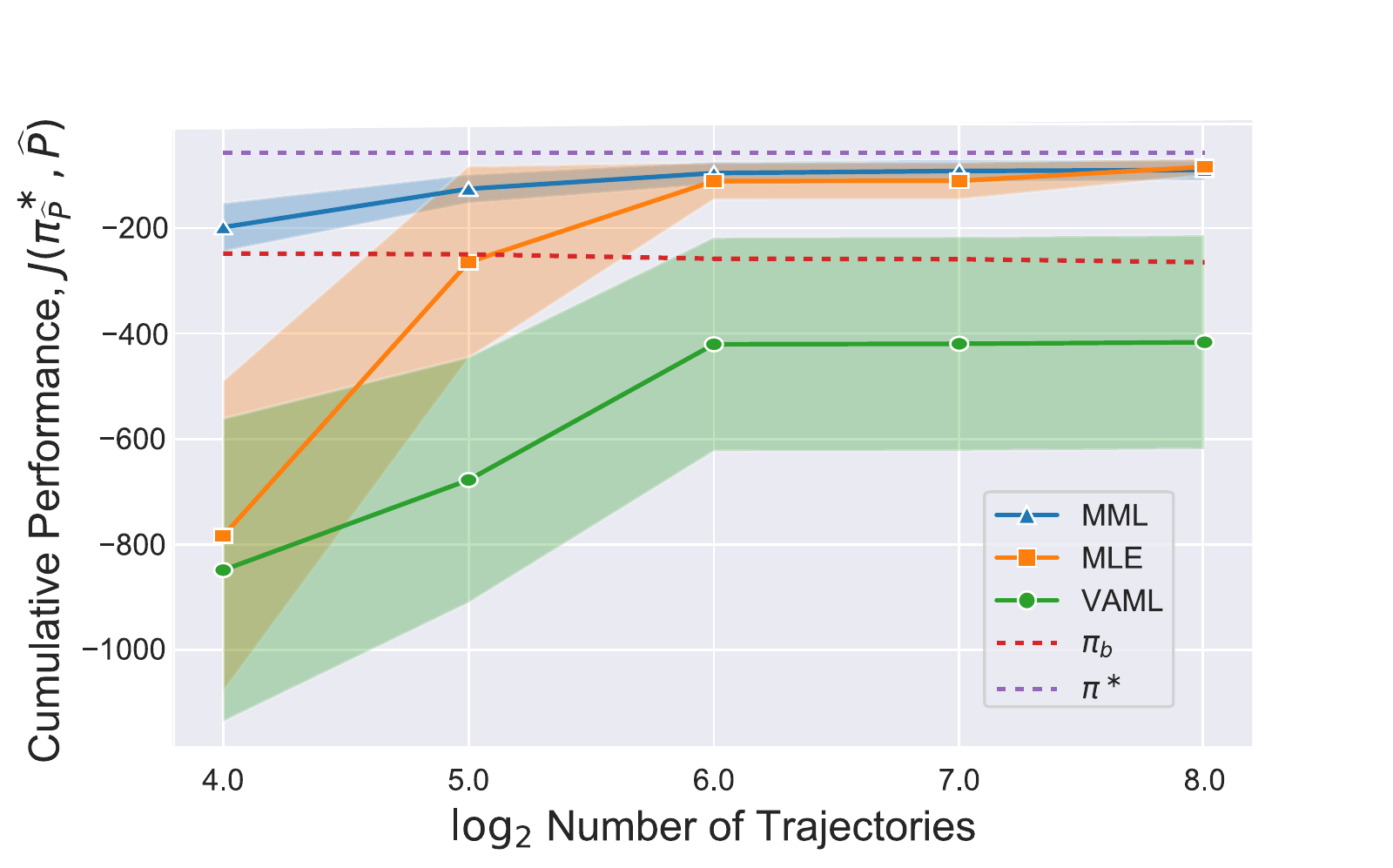}
\vspace{-0.3in}
\caption{\textit{ (Invert. Pend., OPO Performance) Comparison of model-based approaches for OPO with function-approx using Algorithm \ref{algo:main_morel}. Higher is better.} \textit{MML performs competitively even in low data regimes.
}}
 \label{fig:OPO}
 \vspace{-.1in}
\end{figure}

\textbf{Does our approach complement modern offline RL approaches?} We integrate MML, VAML, and MLE with MOREL as in Algorithm \ref{algo:main_morel}. Consequently, Figure \ref{fig:OPO} shows that MML performs competitively with the other methods, achieving near-optimal performance as the number of trajectories increases. MML has good performance even in the low-data regime, whereas other methods perform worse than $\pi_b$. Performance in the low-data regime is of particular interest since sample efficiency is highly desirable.

Algorithm \ref{algo:main_morel} forms a pessimistic MDP where a policy is penalized if it enters a state where there is disagreement between $P_1,\ldots,P_4$. Given that MML performs well in low-data, we can reason that MML produces models with support that stays within the dataset $D$ or generalize well slightly outside this set. The other models poor performance is suggestive of incorrect over-confidence outside of $D$ and PPO produces a policy which takes advantage of this.

\section{Other Related Work}

\textbf{Minimax and Model-Based RL.}
\cite{rajeswaran2020gametheory} introduce an iterative minimax approach to simultaneously find the optimal-policy and a model of the environment.
Despite distribution-shift correction, online data collection is required and is not comparable to MML, where we focus on the batch setting.

\textbf{Batch (Offline) Model-Based RL}
Recent improvements in batch model-based RL focus primarily on the issue of policies taking advantage of errors in the model \citep[]{kidambi2020morel, DeisenrothPILCO2011, Chua2018PETS, Jenner2019MBPO}. 
These improvements typically involve uncertainty quantification to keep the agent in highly certain states to avoid 
model exploitation. 
These improvements are independent of the loss function involved.

\section{Discussion and Future Work}

We have presented a novel approach to learning a model for batch, off-policy model-based reinforcement learning. Our approach follows naturally from the definitions of the OPE and OPO objectives and enjoys distributional robustness and decision-awareness. We examined different scenarios under which our method coincided with other methods as well as when closed form solutions were available. We provided sample complexity analysis and misspecification analysis.  Finally, we empirically validated that our method was competitive with current model learning approaches.

A key component throughout this paper has been the function class $\W \times \V$. Finding other interpretations for this term may prove to be useful outside of MML and is of interest in future work. Furthermore, MML remains part of a two-step OPO pipeline: first learn the model, then return the optimal policy in that model. Another direction of future research is to have a single-shot batch OPO objective that returns both a model and the optimal policy simultaneously, in effect combining MML with the minimax algorithm in \cite{rajeswaran2020gametheory}.
Finally, it may be interesting to integrate MML with other forms of distributionally robust model learning, e.g.,  \citet{liu2020robust}.


\subsubsection*{Acknowledgements}
Cameron Voloshin is supported in part by a Kortschak Fellowship.  This work is also supported in part by NSF \# 1645832, NSF \# 1918839, and funding from Beyond Limits. Nan Jiang is sponsored in part by the DEVCOM Army Research Laboratory under Cooperative Agreement W911NF-17-2-0196 (ARL IoBT CRA). The views and conclusions contained in this document are those of the authors and should not be interpreted as representing the official policies, either expressed or implied, of the Army Research Laboratory or the U.S.~Government. The U.S.~Government is authorized to reproduce and distribute reprints for Government purposes notwithstanding any copyright notation herein.


\bibliography{icml,ci}

\begin{thebibliography}{32}
\providecommand{\natexlab}[1]{#1}
\providecommand{\url}[1]{\texttt{#1}}
\expandafter\ifx\csname urlstyle\endcsname\relax
  \providecommand{\doi}[1]{doi: #1}\else
  \providecommand{\doi}{doi: \begingroup \urlstyle{rm}\Url}\fi

\bibitem[Abachi et~al.(2020)Abachi, Ghavamzadeh, and massoud
  Farahmand]{abachi2020PAML}
Abachi, R., Ghavamzadeh, M., and massoud Farahmand, A.
\newblock Policy-aware model learning for policy gradient methods, 2020.

\bibitem[Antos et~al.(2008)Antos, Szepesv{\'a}ri, and Munos]{antos2008learning}
Antos, A., Szepesv{\'a}ri, C., and Munos, R.
\newblock Learning near-optimal policies with bellman-residual minimization
  based fitted policy iteration and a single sample path.
\newblock \emph{Machine Learning}, 71\penalty0 (1):\penalty0 89--129, 2008.

\bibitem[Bartlett \& Mendelson(2001)Bartlett and
  Mendelson]{bartlett2001Rademacher}
Bartlett, P.~L. and Mendelson, S.
\newblock Rademacher and gaussian complexities: Risk bounds and structural
  results.
\newblock In \emph{Proceedings of the 14th Annual Conference on Computational
  Learning Theory and and 5th European Conference on Computational Learning
  Theory}, Berlin, Heidelberg, 2001. Springer-Verlag.
\newblock ISBN 3540423435.

\bibitem[Bertsekas et~al.(2005)Bertsekas, Bertsekas, Bertsekas, and
  Bertsekas]{bertsekas2005dynamic}
Bertsekas, D.~P., Bertsekas, D.~P., Bertsekas, D.~P., and Bertsekas, D.~P.
\newblock \emph{Dynamic programming and optimal control}, volume~1.
\newblock Athena scientific Belmont, MA, 2005.

\bibitem[Brockman et~al.(2016)Brockman, Cheung, Pettersson, Schneider,
  Schulman, Tang, and Zaremba]{OpenAIgym}
Brockman, G., Cheung, V., Pettersson, L., Schneider, J., Schulman, J., Tang,
  J., and Zaremba, W.
\newblock Openai gym.
\newblock \emph{CoRR}, abs/1606.01540, 2016.

\bibitem[Chen \& Jiang(2019)Chen and Jiang]{chen2019FQI}
Chen, J. and Jiang, N.
\newblock Information-theoretic considerations in batch reinforcement learning.
\newblock In Chaudhuri, K. and Salakhutdinov, R. (eds.), \emph{Proceedings of
  the 36th International Conference on Machine Learning}, Long Beach,
  California, USA, 09--15 Jun 2019. PMLR.

\bibitem[Chua et~al.(2018)Chua, Calandra, McAllister, and Levine]{Chua2018PETS}
Chua, K., Calandra, R., McAllister, R., and Levine, S.
\newblock Deep reinforcement learning in a handful of trials using
  probabilistic dynamics models.
\newblock In Bengio, S., Wallach, H., Larochelle, H., Grauman, K.,
  Cesa-Bianchi, N., and Garnett, R. (eds.), \emph{Advances in Neural
  Information Processing Systems 31}. Curran Associates, Inc., 2018.

\bibitem[Clavera et~al.(2018)Clavera, Rothfuss, Schulman, Fujita, Asfour, and
  Abbeel]{Clavera2018MBMPO}
Clavera, I., Rothfuss, J., Schulman, J., Fujita, Y., Asfour, T., and Abbeel, P.
\newblock Model-based reinforcement learning via meta-policy optimization.
\newblock In \emph{2nd Annual Conference on Robot Learning, CoRL 2018,
  Z{\"{u}}rich, Switzerland, 29-31 October 2018, Proceedings}. {PMLR}, 2018.

\bibitem[Deisenroth \& Rasmussen(2011)Deisenroth and
  Rasmussen]{DeisenrothPILCO2011}
Deisenroth, M.~P. and Rasmussen, C.~E.
\newblock Pilco: A model-based and data-efficient approach to policy search.
\newblock In \emph{Proceedings of the 28th International Conference on
  International Conference on Machine Learning}, Madison, WI, USA, 2011.
  Omnipress.
\newblock ISBN 9781450306195.

\bibitem[Dorobantu \& Taylor(2020)Dorobantu and Taylor]{laypy}
Dorobantu, V. and Taylor, A.
\newblock Lyapy.
\newblock \url{https://github.com/vdorobantu/lyapy}, 2020.

\bibitem[Ernst et~al.(2005)Ernst, Geurts, and Wehenkel]{Ernst2005FQI}
Ernst, D., Geurts, P., and Wehenkel, L.
\newblock Tree-based batch mode reinforcement learning.
\newblock \emph{J. Mach. Learn. Res.}, 6:\penalty0 503–556, December 2005.
\newblock ISSN 1532-4435.

\bibitem[Farahmand(2018)]{farahmand2018IVAML}
Farahmand, A.-m.
\newblock Iterative value-aware model learning.
\newblock In Bengio, S., Wallach, H., Larochelle, H., Grauman, K.,
  Cesa-Bianchi, N., and Garnett, R. (eds.), \emph{Advances in Neural
  Information Processing Systems 31},  9072--9083. Curran Associates, Inc.,
  2018.

\bibitem[Farahmand et~al.(2017)Farahmand, Barreto, and
  Nikovski]{farahmand2017VAML}
Farahmand, A.-M., Barreto, A., and Nikovski, D.
\newblock {Value-Aware Loss Function for Model-based Reinforcement Learning}.
\newblock In Singh, A. and Zhu, J. (eds.), \emph{Proceedings of the 20th
  International Conference on Artificial Intelligence and Statistics}, Fort
  Lauderdale, FL, USA, 20--22 Apr 2017. PMLR.

\bibitem[Feng et~al.(2019)Feng, Li, and Liu]{feng2019kernel}
Feng, Y., Li, L., and Liu, Q.
\newblock A kernel loss for solving the bellman equation.
\newblock In \emph{Advances in Neural Information Processing Systems}, 2019.

\bibitem[Goodfellow et~al.()Goodfellow, Bengio, and
  Courville]{Goodfellow2016DL}
Goodfellow, I., Bengio, Y., and Courville, A.
\newblock \emph{Deep Learning}.
\newblock MIT Press.

\bibitem[Goodfellow et~al.(2014)Goodfellow, Pouget-Abadie, Mirza, Xu,
  Warde-Farley, Ozair, Courville, and Bengio]{Goodfellow2014GAN}
Goodfellow, I., Pouget-Abadie, J., Mirza, M., Xu, B., Warde-Farley, D., Ozair,
  S., Courville, A., and Bengio, Y.
\newblock Generative adversarial nets.
\newblock In Ghahramani, Z., Welling, M., Cortes, C., Lawrence, N.~D., and
  Weinberger, K.~Q. (eds.), \emph{Advances in Neural Information Processing
  Systems 27},  2672--2680. Curran Associates, Inc., 2014.

\bibitem[Janner et~al.(2019)Janner, Fu, Zhang, and Levine]{Jenner2019MBPO}
Janner, M., Fu, J., Zhang, M., and Levine, S.
\newblock When to trust your model: Model-based policy optimization.
\newblock In Wallach, H., Larochelle, H., Beygelzimer, A., d\textquotesingle
  Alch\'{e}-Buc, F., Fox, E., and Garnett, R. (eds.), \emph{Advances in Neural
  Information Processing Systems 32},  12519--12530. Curran Associates, Inc.,
  2019.

\bibitem[Kidambi et~al.(2020)Kidambi, Rajeswaran, Netrapalli, and
  Joachims]{kidambi2020morel}
Kidambi, R., Rajeswaran, A., Netrapalli, P., and Joachims, T.
\newblock Morel : Model-based offline reinforcement learning, 2020.

\bibitem[Kingma \& Ba(2015)Kingma and Ba]{kingma2014Adam}
Kingma, D.~P. and Ba, J.
\newblock Adam: {A} method for stochastic optimization.
\newblock In Bengio, Y. and LeCun, Y. (eds.), \emph{3rd International
  Conference on Learning Representations, {ICLR} 2015, San Diego, CA, USA, May
  7-9, 2015, Conference Track Proceedings}, 2015.

\bibitem[Kurutach et~al.(2018)Kurutach, Clavera, Duan, Tamar, and
  Abbeel]{kurutach2018metrpo}
Kurutach, T., Clavera, I., Duan, Y., Tamar, A., and Abbeel, P.
\newblock Model-ensemble trust-region policy optimization.
\newblock In \emph{6th International Conference on Learning Representations,
  {ICLR} 2018, Vancouver, BC, Canada, April 30 - May 3, 2018, Conference Track
  Proceedings}. OpenReview.net, 2018.

\bibitem[Liu et~al.(2020)Liu, Shi, Chung, Anandkumar, and Yue]{liu2020robust}
Liu, A., Shi, G., Chung, S.-J., Anandkumar, A., and Yue, Y.
\newblock Robust regression for safe exploration in control.
\newblock In \emph{Learning for Dynamics and Control (L4DC)}, 2020.

\bibitem[Liu et~al.(2018)Liu, Li, Tang, and Zhou]{liu2018breaking}
Liu, Q., Li, L., Tang, Z., and Zhou, D.
\newblock Breaking the curse of horizon: Infinite-horizon off-policy
  estimation.
\newblock In \emph{Advances in Neural Information Processing Systems}, 2018.

\bibitem[Luo et~al.(2019)Luo, Xu, Li, Tian, Darrell, and Ma]{Luo2019SLBO}
Luo, Y., Xu, H., Li, Y., Tian, Y., Darrell, T., and Ma, T.
\newblock Algorithmic framework for model-based deep reinforcement learning
  with theoretical guarantees.
\newblock In \emph{7th International Conference on Learning Representations,
  {ICLR} 2019, New Orleans, LA, USA, May 6-9, 2019}. OpenReview.net, 2019.

\bibitem[MacKay(2002)]{MacKay2002MCMC}
MacKay, D. J.~C.
\newblock \emph{Information Theory, Inference \& Learning Algorithms}.
\newblock Cambridge University Press, USA, 2002.
\newblock ISBN 0521642981.

\bibitem[Mohri et~al.(2012)Mohri, Rostamizadeh, and
  Talwalkar]{mohri2012foundations}
Mohri, M., Rostamizadeh, A., and Talwalkar, A.
\newblock \emph{Foundations of machine learning}.
\newblock MIT press, 2012.

\bibitem[Raffin et~al.(2019)Raffin, Hill, Ernestus, Gleave, Kanervisto, and
  Dormann]{stablebaselines3}
Raffin, A., Hill, A., Ernestus, M., Gleave, A., Kanervisto, A., and Dormann, N.
\newblock Stable baselines3.
\newblock \url{https://github.com/DLR-RM/stable-baselines3}, 2019.

\bibitem[Rajeswaran et~al.(2020)Rajeswaran, Mordatch, and
  Kumar]{rajeswaran2020gametheory}
Rajeswaran, A., Mordatch, I., and Kumar, V.
\newblock A game theoretic framework for model based reinforcement learning,
  2020.

\bibitem[Schaefer \& Anandkumar(2019)Schaefer and Anandkumar]{Schaefer2019CGD}
Schaefer, F. and Anandkumar, A.
\newblock Competitive gradient descent.
\newblock In Wallach, H., Larochelle, H., Beygelzimer, A., d\textquotesingle
  Alch\'{e}-Buc, F., Fox, E., and Garnett, R. (eds.), \emph{Advances in Neural
  Information Processing Systems 32},  7625--7635. Curran Associates, Inc.,
  2019.

\bibitem[Schulman et~al.(2017)Schulman, Wolski, Dhariwal, Radford, and
  Klimov]{Schulman17PPO}
Schulman, J., Wolski, F., Dhariwal, P., Radford, A., and Klimov, O.
\newblock Proximal policy optimization algorithms.
\newblock \emph{CoRR}, abs/1707.06347, 2017.

\bibitem[Sutton(1990)]{Sutton1990Dyna}
Sutton, R.~S.
\newblock Integrated architectures for learning, planning, and reacting based
  on approximating dynamic programming.
\newblock In \emph{In Proceedings of the Seventh International Conference on
  Machine Learning}. Morgan Kaufmann, 1990.

\bibitem[Uehara et~al.(2020)Uehara, Huang, and Jiang]{uehara2019minimax}
Uehara, M., Huang, J., and Jiang, N.
\newblock {Minimax Weight and Q-Function Learning for Off-Policy Evaluation}.
\newblock In \emph{Proceedings of the 37th International Conference on Machine
  Learning}, 2020.

\bibitem[Voloshin et~al.(2019)Voloshin, Le, Jiang, and
  Yue]{voloshin2019empirical}
Voloshin, C., Le, H.~M., Jiang, N., and Yue, Y.
\newblock Empirical study of off-policy policy evaluation for reinforcement
  learning.
\newblock \emph{arXiv preprint arXiv:1911.06854}, 2019.

\end{thebibliography}
\bibliographystyle{clv}
\nocite{voloshin2019empirical}
\clearpage

\tableofcontents
\onecolumn
\appendix
\section{Glossary of Terms}
\label{app:glossary}

\begin{figure}[!h]
\begin{minipage}[t]{1\textwidth}
\vspace{-.1in}
\captionof{table}{Glossary of terms}
\label{tab:glossary}
\vspace{-0.05in}
\begin{center}
\begin{small}
\begin{tabular}{ll}
\toprule
 Acronym & Term \\
 \midrule
OPE & Off Policy (Policy) Evaluation \\
OPO & Off Policy (Policy) Optimization. Also goes by batch off-policy reinforcement learning. \\
$\State$ & State Space \\
$\mathcal{A}$ & Action Space \\
$P$ & Transition Function \\
$P^\ast$ & True Transition Function \\
$\mathcal{R}$ & Reward Function \\
$\mathcal{X}$ & State-Action Space $\State \times \mathcal{A}$ \\
$\gamma$ & Discount Factor \\
$\pi$ & Policy \\
$J(\pi, P)$ & Performance of $\pi$ in $P$ \\
$V^P_{\pi}$ & Value Function of $\pi$ with respect to $P$ \\
$d_0$ & Initial State Distribution \\
$d_\pi^{P, \gamma}$ & (Discounted) Distribution of State-Action Pairs Induced by Running $\pi$ in $P$ \\
$w^P_{\pi}$ & Distribution Shift ($w^P_{\pi}(s,a) = \frac{d_\pi^{P, \gamma}(s,a)}{D_{\pi_b}(s,a)}$) \\
$\nu$ & Lebesgue measure \\
$d_{\pi_b}$ & Behavior state distribution \\
$\pi_b$ & Behavior policy \\
$D_{\pi_b}$ & Behavior data ($d_{\pi_b} \pi_b$) \\
$D$ & Dataset containing samples from $D_{\pi_b} P^\ast$ \\
$E_n[\cdot]$ & Empirical approximation using $D$ \\
$E[\cdot]$ & Exact expectation \\
$\W$ & Distribution Shifts Function Class (e.g. $\frac{d_\pi^{P}(s,a)}{D_\pi(s,a)}$) \\
$\V$ & Value Function Class (e.g. $V^P_\pi \in \V)$ \\
$\M$ & Model Function Class (e.g. $P \in \M)$ \\
$\Loss$ & Model Learning Loss Function \\
$\widehat{P}$ & Best Model w.r.t $\Loss$ \\
$\epsilon_\Hh$ & Misspecification Error \\
$\pi^\ast_P$ & Optimal Policy in $P$ \\
RKHS & Reproducing Kernel Hilbert Space \\
LQR & Linear Quadratic Regulator \\
IP & Inverted Pendulum \\
MML & Minimax Model Learning (Ours) \\
MLE & Maximum Likelihood Estimation \\
VAML & Value-Aware Model Learning
\end{tabular}
\end{small}
\end{center}

  \end{minipage}
  \end{figure}
  
 \clearpage
\section{OPE}

In this section we explore the OPE results in the order in which they were presented in the main paper. 

\subsection{Main Result}
\label{app:ope:sec:main}


\begin{proof}
[Proof for Theorem \ref{thm:opeerror}]
Assume $(w^{P^\ast}_\pi, \val{\pi}{P}) \in \W \times \V$. Fix some $P \in \M$. We use both definitions of $J$ as follows
\begin{align*}
J(\pi, P) - J(\pi, P^\ast) &= E_{d_0}[\val{\pi}{P}] - E_{(s,a) \sim \density{\pi}{P^\ast}, r \sim \mathcal{R}(\cdot|s,a)}[r] \\
&= E_{(s,a) \sim \density{\pi}{P^\ast}}[\val{\pi}{P}(s) - E_{r\sim \mathcal{R}(\cdot|s,a)}[r]] + E_{d_0}[\val{\pi}{P}] -  E_{(s,a) \sim \density{\pi}{P^\ast}}[\val{\pi}{P}(s)] \\
&= E_{(s,a) \sim \density{\pi}{P^\ast}}[\val{\pi}{P}(s) - E_{r\sim \mathcal{R}(\cdot|s,a)}[r]] -  \sum_{t=1}^\infty \gamma^t \int \densityt{\pi}{P^\ast}{t}(s,a) \val{\pi}{P}(s) d\nu(s,a) \\
&= E_{(s,a) \sim \density{\pi}{P^\ast}}[\gamma E_{s'\sim P(\cdot|s,a)}[\val{\pi}{P}(s')]] -  \gamma \sum_{t=0}^\infty \gamma^t \int \densityt{\pi}{P^\ast}{t+1}(s,a) \val{\pi}{P}(s) d\nu(s,a) \\
&= \gamma E_{(s,a) \sim \density{\pi}{P^\ast}}[ E_{s'\sim P(\cdot|s,a)}[\val{\pi}{P}(s')]] - \gamma \sum_{t=0}^\infty \gamma^t \int \densityt{\pi}{P^\ast}{t}(\tilde{s},\tilde{a}) P^\ast(s|\tilde{s},\tilde{a}) \pi(a|s) \val{\pi}{P}(s) d\nu(\tilde{s},\tilde{a},s,a) \\
&= \gamma E_{(s,a) \sim \density{\pi}{P^\ast}}[ E_{s'\sim P(\cdot|s,a)}[\val{\pi}{P}(s')]] - \gamma \sum_{t=0}^\infty \gamma^t \int \densityt{\pi}{P^\ast}{t}(s,a) P^\ast(s'|s,a) \val{\pi}{P}(s') d\nu(s,a,s') \\
&= \gamma E_{(s,a) \sim \density{\pi}{P^\ast}}[ E_{s'\sim P(\cdot|s,a)}[\val{\pi}{P}(s')]] - \gamma E_{(s,a) \sim \density{\pi}{P^\ast}}[ E_{s'\sim P^\ast(\cdot|s,a)}[\val{\pi}{P}(s')]] \\
&= \gamma E_{(s,a) \sim \density{\pi}{P^\ast}}[ E_{s'\sim P(\cdot|s,a)}[\val{\pi}{P}(s')] - E_{s'\sim P^\ast(\cdot|s,a)}[\val{\pi}{P}(s')]] \\
&= \gamma E_{(s,a,s') \sim D_{\pi_b} P^\ast(\cdot|s,a)}[ \frac{\density{\pi}{P^\ast}(s,a)}{D_{\pi_b}(s,a)} \left( E_{x\sim P(\cdot|s,a)}[\val{\pi}{P}(x)] - \val{\pi}{P}(s') \right)]] \\
&= \gamma E_{(s,a,s') \sim D_{\pi_b} P^\ast(\cdot|s,a)}[ w^{P^\ast}_\pi(s,a) \left( E_{x\sim P(\cdot|s,a)}[\val{\pi}{P}(x)] - \val{\pi}{P}(s') \right)]] \\
&= \gamma \Loss(w^{P^\ast}_\pi, \val{\pi}{P}, P),
\end{align*}
where the first equality is definition. The second equality is addition of $0$. The third equality is simplification. The fourth equality is change of bounds. The fifth is definition. The sixth is relabeling of the integration variables. The seventh and eighth are simplification. The ninth is importance sampling. The tenth and last is definition.
Since $(w^{P^\ast}_\pi, \val{\pi}{P}) \in \W \times \V$ then 
$$ |J(\pi, P) - J(\pi, P^\ast)| = \gamma |\Loss(w^{P^\ast}_\pi, \val{\pi}{P}, P)| \leq \gamma \max_{w \in \W, V \in \V} |\Loss(w, V, P)| \leq \gamma \min_{P \in \M} \max_{w \in \W, V \in \V} |\Loss(w, V, P)|, $$
where the last inequality holds because $P$ was selected in $\M$ arbitrarily.

Now, instead, assume $(w^{P}_\pi, \val{\pi}{P^\ast}) \in \W \times \V$. Fix some $P \in \M$. Then, similarly,
\begin{align*}
J(\pi, P) - J(\pi, P^\ast) &= E_{(s,a) \sim \density{\pi}{P}, r \sim \mathcal{R}(\cdot|s,a)}[r] - E_{d_0}[\val{\pi}{P^\ast}] \\
&= E_{(s,a) \sim \density{\pi}{P}}[\val{\pi}{P^\ast}(s)] - E_{d_0}[\val{\pi}{P^\ast}]  - E_{(s,a) \sim \density{\pi}{P}}[\val{\pi}{P^\ast}(s) - E_{r\sim \mathcal{R}(\cdot|s,a)}[r]] \\
&= \sum_{t=1}^\infty \gamma^t \int \densityt{\pi}{P}{t}(s,a) \val{\pi}{P^\ast}(s) d\nu(s,a) - E_{(s,a) \sim \density{\pi}{P}}[\val{\pi}{P^\ast}(s) - E_{r\sim \mathcal{R}(\cdot|s,a)}[r]] \\
&= \gamma \sum_{t=0}^\infty \gamma^t \int \densityt{\pi}{P}{t+1}(s,a) \val{\pi}{P^\ast}(s) d\nu(s,a) - E_{(s,a) \sim \density{\pi}{P}}[\gamma E_{s'\sim P^\ast(\cdot|s,a)}[\val{\pi}{P^\ast}(s')]] \\
&= \gamma \sum_{t=0}^\infty \gamma^t \int \densityt{\pi}{P}{t}(\tilde{s},\tilde{a}) P(s|\tilde{s},\tilde{a}) \pi(a|s) \val{\pi}{P^\ast}(s) d\nu(\tilde{s},\tilde{a},s,a) - \gamma E_{(s,a) \sim \density{\pi}{P}}[ E_{s'\sim P^\ast(\cdot|s,a)}[\val{\pi}{P^\ast}(s')]] \\
\end{align*}
\begin{align*}
&= \gamma \sum_{t=0}^\infty \gamma^t \int \densityt{\pi}{P}{t}(s,a) P(s'|s,a) \val{\pi}{P^\ast}(s') d\nu(s,a,s') - \gamma E_{(s,a) \sim \density{\pi}{P}}[ E_{s'\sim P^\ast(\cdot|s,a)}[\val{\pi}{P^\ast}(s')]] \\
&= \gamma E_{(s,a) \sim \density{\pi}{P}}[ E_{s'\sim P(\cdot|s,a)}[\val{\pi}{P^\ast}(s')]] - \gamma E_{(s,a) \sim \density{\pi}{P}}[ E_{s'\sim P^\ast(\cdot|s,a)}[\val{\pi}{P^\ast}(s')]] \\
&= \gamma E_{(s,a) \sim \density{\pi}{P}}[ E_{s'\sim P(\cdot|s,a)}[\val{\pi}{P^\ast}(s')] - E_{s'\sim P^\ast(\cdot|s,a)}[\val{\pi}{P^\ast}(s')]] \\
&= \gamma E_{(s,a,s') \sim D_{\pi_b} P^\ast(\cdot|s,a)}[ \frac{\density{\pi}{P}(s,a)}{D_{\pi_b}(s,a)} \left( E_{x\sim P(\cdot|s,a)}[\val{\pi}{P^\ast}(x)] - \val{\pi}{P^\ast}(s') \right)]] \\
&= \gamma E_{(s,a,s') \sim D_{\pi_b} P^\ast(\cdot|s,a)}[ w^{P}_\pi(s,a) \left( E_{x\sim P(\cdot|s,a)}[\val{\pi}{P^\ast}(x)] - \val{\pi}{P^\ast}(s') \right)]] \\
&= \gamma \Loss(w^{P}_\pi, \val{\pi}{P^\ast}, P),
\end{align*}
where we follow the same steps as in the previous derivation. Since $(w^{P}_\pi, \val{\pi}{P^\ast}) \in \W \times \V$ then
$$ |J(\pi, P) - J(\pi, P^\ast)| = \gamma |\Loss(w^{P}_\pi, \val{\pi}{P^\ast}, P)| \leq \gamma \max_{w \in \W, V \in \V} |\Loss(w, V, P)| \leq \gamma \min_{P \in \M} \max_{w \in \W, V \in \V} |\Loss(w, V, P)|, $$
where the last inequality holds because $P$ was selected in $\M$ arbitrarily.
\end{proof}


\subsection{Sample Complexity for OPE}\label{app:ope:sec:samplecomplexity}

We do not have access to exact expectations, so we must work with $\widehat{P}_n = \arg\min_{P}\max_{w,V} E_n[\ldots]$ instead of $\widehat{P} = \arg\min_{P}\max_{w,V} E[\ldots]$. Furthermore, $J(\pi, \widehat{P})$ requires exact expectation of an infinite sum: $E_{d_0}[\sum_{t=0}^\infty \gamma^t r_t]$ where we collect $r_t$ by running $\pi$ in simulation $\widehat{P}$. Instead, we can only estimate an empirical average over a finite sum in $\widehat{P}_n$: $J_{T,m}(\pi, \widehat{P}_n) = \frac{1}{m} \sum_{j=1}^m \sum_{t=0}^T \gamma^t r^j_t$, where each $j$ indexes rollouts starting from $s_0 \sim d_0$ and the simulation is over $\widehat{P}_n$. Our OPE estimate is therefore bounded as follows:

\begin{thm}\label{app:ope:thm:samplecomplexity}[OPE Error] Let the functions in $\V$ and $\W$ be uniformly bounded by $C_\V$ and $C_\W$ respectively. Assume the conditions of Theorem \ref{thm:opeerror} hold and $|\mathcal{R}| \leq \Rmax, \gamma \in [0,1)$. Then, with probability $1-\delta$, 
\begin{align*}
|&J_{T,m}(\pi,\widehat{P}_n) - J(\pi,P^\ast)| \leq \gamma \min_{P}\max_{w,V} |\Loss(w, V, P)| \\& + 4 \gamma \mathfrak{R}_n(\W, \V, \M) + \frac{2  \Rmax}{1-\gamma} \gamma^{T+1} \\
& + \frac{2 \Rmax}{1-\gamma}\sqrt{\log(2/\delta)/(2m)} + 4 \gamma C_\W C_\V \sqrt{\log(2/\delta)/n}
\end{align*}
where $\mathfrak{R}_n(\W, \V, \M)$ is the Rademacher complexity of the function class 
\begin{align*}
\{(s,a,s') \mapsto   &w(s,a)(E_{x\sim P}[V(x)] - V(s')) : \\ &w \in \mathcal{W}, V \in \mathcal{V}, P \in \mathcal{\M}\}.
\end{align*}
\end{thm}


\begin{proof}
[Proof for Theorem \ref{app:ope:thm:samplecomplexity}] 

By definition and triangle inequality,
\begin{Salign}
|J_{T,m}(\pi,\widehat{P}_n ) - J(\pi,P^\ast)| &= | J_{T,m}(\pi,\widehat{P}_n ) - J(\pi,\widehat{P}_n) + J(\pi,\widehat{P}_n) - J(\pi,P^\ast)| \quad  \\
&\leq  \underbrace{| J_{T,m}(\pi,\widehat{P}_n ) - J(\pi,\widehat{P}_n )|}_{(a)} + \underbrace{|J(\pi,\widehat{P}_n ) - J(\pi,P^\ast)|}_{(b)} 
\end{Salign}

Define $\widehat{V}_{\pi, T}^P(s_0^i) \equiv \sum_{t=0}^T \gamma^t r_t^i$ for some trajectory indexed by $i \in \mathbb{N}$ where $r_t^i$ is the reward obtained by running $\pi$ in $P$ at time $t \leq T$ starting at $s_0^i$. For $(a)$,
\begin{Salign}
| J_{T,m}(\pi,\widehat{P}_n ) - J(\pi,\widehat{P}_n )| &= \left| \frac{1}{m} \sum_{i=1}^m \widehat{V}^{\widehat{P}_n }_{\pi, T} (s_0^i) - \frac{1}{m} \sum_{i=1}^m \widehat{V}^{\widehat{P}_n }_{\pi, \infty} (s_0^i) + \frac{1}{m} \sum_{i=1}^m \widehat{V}^{\widehat{P}_n }_{\pi, \infty} (s_0^i) - E_{d_0}[V^{\widehat{P}_n }_{\pi}] \right| \\
&\leq \left| \frac{1}{m} \sum_{i=1}^m \widehat{V}^{\widehat{P}_n }_{\pi, T} (s_0^i) - \frac{1}{m} \sum_{i=1}^m \widehat{V}^{\widehat{P}_n }_{\pi, \infty} (s_0^i)\right| + 
\left|\frac{1}{m} \sum_{i=1}^m \widehat{V}^{\widehat{P}_n }_{\pi, \infty} (s_0^i) - E_{d_0}[V^{\widehat{P}_n }_{\pi}] \right| \\
&\leq \frac{ 2 \Rmax}{1-\gamma} \gamma^{T+1} + \frac{ 2 \Rmax}{1-\gamma}\sqrt{\log(2/\delta)/(2m)},
\end{Salign}
with probability $1-\delta$, where the last inequality is definition of $\widehat{V}_{\pi,T}$ and Hoeffding's inequality. 

For $(b)$, by Theorem \ref{thm:opeerror},
\begin{Salign}
|J&(\pi,\widehat{P}_n ) - J(\pi,P^\ast)| \\
&= \gamma |L(w_{\pi}^{P^\ast}, V^{\widehat{P}_n}, \widehat{P}_n)|  \\
&\leq \gamma \max_{w,V} |L(w, V, \widehat{P}_n)| \\
&= \gamma(\max_{w,V} |L(w, V, \widehat{P}_n)| - \max_{w,V} |L_n(w, V, \widehat{P}_n)| + \max_{w,V} |L_n(w, V, \widehat{P}_n)| - \max_{w,V} |L(w, V, \widehat{P})|  + \max_{w,V} |L(w, V, \widehat{P})|) \\
&\leq \gamma( 2 \max_{w, V, P}\left| |L(w, V, P)| -  |L_n(w, V, P)|\right| + \min_{P}\max_{w,V} |L(w, V, P)|) \\
&\leq \gamma (2 \mathfrak{R}'_n(\W,\V,\M) + 2 K \sqrt{\log(2/\delta)/n} +  \min_{P}\max_{w,V} |L(w, V, P)| ) \\
&\leq \gamma(4\mathfrak{R}_n(\W,\V,\M) + 2 K \sqrt{\log(2/\delta)/n} +  \min_{P}\max_{w,V} |L(w, V, P)|)
\end{Salign}
where $\mathfrak{R}'_n(\W, \V,\M)$ is the Rademacher complexity of the function class 
$$
\{(s,a,s') \mapsto  | w(s,a)(E_{x\sim P}[V(x)] - V(s'))| : w \in \W, V \in \V, P \in \M \}
$$
noting that $K = 2 C_w C_V $ uniformly bounds $| w(s,a)(E_{x \sim P(\cdot|s,a)}[V(x)] - V(s'))|$ (Theorem 8 \cite{bartlett2001Rademacher}). Furthermore since absolute value is $1$-Lipshitz (by reverse triangle ineq), then $\mathfrak{R}'_n < 2 \mathfrak{R}_n$ (Theorem 12 \cite{bartlett2001Rademacher}) where $\mathfrak{R}_n(\W, \V, \M)$ is the Rademacher complexity of the function class 
$$
\{(s,a,s') \mapsto   w(s,a)(E_{x \sim P(\cdot|s,a)}[V(x)] - V(s'))) : w \in \W, V \in \V, P \in \M \}.
$$
Altogether, combining (1), (2), (3) we get our result.
\end{proof}

The first term can be thought of as the estimate under infinite data, the second term as the penalty for using function classes that are too rich, and the remaining terms as the price we pay for finite data/ finite calculations.

\subsection{Misspecification for OPE}
\label{app:ope:sec:misspec}


When the assumptions behind MML do not hold, our method underbounds the true error. The following is the proof for this Proposition.

\begin{proof}
[Proof for Prop. \ref{lem:misspec}]
We have shown already that $J(\pi, \widehat{P}) - J(\pi, P^\ast) = \gamma \Loss(w^{P^\ast}_{\pi},V^P_{\pi},P) \;(= \gamma \Loss((WV)^\ast,P))$. Therefore, by linearity of $\Loss$ in $\Hh,$ we have
\begin{align*}
    |\Loss((WV)^\ast, P)| &= |\Loss(h, P) + \Loss((WV)^\ast - h, P)| \quad \forall h \in \Hh, P \in \M\\
    &\leq |\Loss(h, P)| + |\Loss( (WV)^\ast - h, P)| \\
    &\leq \min_P \max_h |\Loss(h,P)| + |\Loss(h - (WV)^\ast, P)| \\
    &\leq \min_P \max_h |\Loss(h,P)| + \max_P \min_h |\Loss((WV)^\ast - h, P)|
\end{align*}
where $\epsilon_\Hh = \max_P \min_h |\Loss((WV)^\ast - h, P)|.$
Therefore $|J(\pi, \widehat{P}) - J(\pi, P^\ast)| \leq \gamma (\min_P \max_h |\Loss(h,P)| + \epsilon_\Hh),$ as desired.
\end{proof}



\subsection{Application to the Online Setting and Brief VAML Comparison}
\label{app:ope:sec:batch2online}

Algorithm \ref{algo:online} is the prototypical online model-based RL algorithm. In contrast to the batch setting, we allow for online data collection. We require a function called PLANNER, which can take a model $P_k$ and find the optimal solution $\pi_k$ in $P_k$.

\begin{algorithm}[H]
	\caption{Online Model-Based RL} 
	\label{algo:online}
	\begin{algorithmic}[1]
    	\REQUIRE $\pi_0 = \pi_b$. PLANNER($\cdot$)
	    \FOR{$k = 0,1,\ldots,K$} 
	        \STATE Collect data $D_k$ by interacting with the true environment using $\pi_k$.
	        \STATE Fit $P_k \leftarrow \arg\min_{P \in \M} \max_{w,V \in \W,\V} \Loss_{MML}(w,V,P)$ where $D_{\pi_b} = D_k$\\
	        \STATE Fit $\pi_{k} \leftarrow \text{PLANNER}(P_k)$
	    \ENDFOR
	\RETURN ($P_K$, $\pi_K$)
	\end{algorithmic}
\end{algorithm}

Here we show that MML lower bounds the VAML error in online model-based RL, where VAML is designed. 

\begin{prop}\label{app:ope:prop:MML_tighter} Let $\W = \{1\}$. Then
$$\min_{P \in \M} \max_{w \in \W,V \in \V} \Loss_{MML}(w,V,P)^2 \leq \min_{P \in \M} \Loss_{VAML}(\V,P), $$ for every $\V, \M$.
\end{prop}
\begin{proof}
Fix $P \in \M$. Then, by definition, $\Loss_{MML}(w,V,P) = E_{(s,a,s') \sim D_{\pi_b} P^\ast}[w(s,a) (E_{x \sim P(\cdot|s,a)}[V(x)] - V(s'))]$. Since $\W = \{1\}$, then we can eliminate this dependence and get $\Loss_{MML}(1,V,P) = E_{(s,a,s') \sim D_{\pi_b} P^\ast}[E_{x \sim P(\cdot|s,a)}[V(x)] - V(s')].$ Explicitly,
\begin{align*}
    \Loss_{MML}(1, V, P)^2 &= (\int \left( \int P(x|s,a) V(x) d\nu(x) - \int P^\ast(s'|s,a) V(s') d\nu(s') \right) d\nu(s,a) )^2  \\
    &= (\int \left( \int (P(x|s,a) - P^\ast(x|s,a)) V(s') d\nu(x) \right) d\nu(s,a) )^2 \\
    &\leq \int \left( \int (P(x|s,a) - P^\ast(x|s,a)) V(x) d\nu(x) \right)^2 d\nu(s,a) , \quad \text{ Cauchy Schwarz} \\
\end{align*}
Taking the $\max_{V \in \V}$ on both sides and noting $\max_{V} \int f(V) \leq \int \max_{V} f(V)$ for any $f, V$ then
\begin{align}\label{eq:MML_VAML_ineq}
    \max_{V \in \V} \Loss_{MML}(1, V, P)^2 &\leq \int \max_{V \in \V}  \left( \int (P(x|s,a) - P^\ast(x|s,a)) V(x) d\nu(x) \right)^2 d\nu(s,a) \\
    &= \Loss_{VAML}(\V, P).
\end{align}
Since we chose $P$ arbitrarily, then Eq \ref{eq:MML_VAML_ineq} holds for any $P \in \M$. In particular, if $\widehat{P}_{VAML} = \arg\min_{P \in \M} \Loss_{VAML}(\V, P)$ then 
$$ \min_{P \in \M} \max_{V \in \V} \Loss_{MML}(1, V, P)^2 \leq \max_{V \in \V} \Loss_{MML}(1, V, \widehat{P}_{VAML})^2 \leq \min_{P \in \M} \Loss_{VAML}(\V, P) $$
\end{proof}

Prop \ref{app:ope:prop:MML_tighter} reflects that the MML loss function is a tighter loss in the online model-based RL case than VAML. In a sense, this reflects that MML should be the preferred decision-aware loss function even in online model-based RL. An argument in favor of VAML is that it is more computationally tractable given an assumption that $\V$ is the set of linear function approximators. However, if we desire to use more powerful function approximation VAML suffers the same computational issues as MML. In general the pointwise supremum within VAML presents a substantial computational challenge while the uniform supremum from MML is much more mild, can be formulated as a two player game and solved via higher-order gradient descent (see Section \ref{app:scenarios:rkhs}). 

Lastly, VAML defines the pointwise loss with respect to the $L^2$ norm of the difference between $P$ and $P^\ast$. The choice is justified in that it is computationally friendlier but it is noted that $L^1$ may also be reasonable \cite[]{farahmand2017VAML}. We show in the following example that, actually, VAML may not work with a pointwise $L^1$ error. 

\begin{exmp}
Let $\State = A \cup B$, a disjoint partition of the state space. For simplicity, assume no dependence on actions. Suppose our models $\M = \{P_\alpha\}_{\alpha \in [0,1]}$ take the form
$$ P_\alpha(s'|s) = \begin{cases}
\alpha &\quad s' \in A \\
1-\alpha &\quad s' \in B \\
\end{cases}
$$
Suppose also that $P^\ast_{\alpha^\ast} \in \M$ for some $\alpha^\ast \in [0,1]$. Let $\V = \{x \mathbf{1}_{s \in A}(s) + y \mathbf{1}_{s \in B}(s)| x,y < M \in \R^+ \}$ be all bounded piecewise constant value functions with $\|V\|_\infty = M \in \R^+$. Then the empirical VAML loss with $L^1$ pointwise distance does not choose $P^\ast$ when $\alpha \neq \frac{1}{2}$ and cannot differentiate between $P^\ast$ and any other $P \in \M$ when $\alpha^\ast = \frac{1}{2}$. MML does not have this issue.

\begin{proof}
To show this, first fix $P \in \M$. Then the empirical VAML loss (in expectation) is given by
\begin{align*}
    E_{s \sim P^\ast}[\max_{V}|E_{x \sim P}[V(x)] - V(s)|] &= \alpha^\ast \max_{V}|E_{x \sim P}[V(x)] - V(A)| + (1-\alpha^\ast) \max_{V}|E_{x \sim P}[V(x)] - V(B)| \\
    &= \alpha^\ast \max_{x,y \in [0,M]}|\alpha x + (1-\alpha) y - x| + (1-\alpha^\ast) \max_{x,y \in [0,M]}|\alpha x + (1-\alpha) y - y| \\
    &= \alpha^\ast \max_{x,y \in [0,M]}|(\alpha-1)(x-y)| + (1-\alpha^\ast) \max_{x,y \in [0,M]}|\alpha(x-y)| \\
    &= (\alpha^\ast |\alpha - 1| + (1-\alpha^\ast) |\alpha|) M
\end{align*}
If $\alpha^\ast < .5$ then the minimizer of the above quantity is $\alpha = 0$, if $\alpha^\ast > .5$ then the minimizer is $\alpha = 1$. Therefore, if $\alpha^\ast \in (0,.5) \cup (.5,1)$ then VAML picks the wrong model $\alpha \neq \alpha^\ast$. Additionally, in the case that $\alpha^{\ast} = .5$ then the loss is $\frac{M}{2}$ for every $P \in \M$. In this case, VAML with $L^1$ cannot differentiate between any model; all models are perfectly identical.

On the other hand, we repeat this process with MML:
\begin{align*}
    |E_{s \sim P^\ast}[E_{x \sim P}[V(x)] - V(s)]| &= |\alpha^\ast (E_{x \sim P}[V(x)] - V(A)) + (1-\alpha^\ast) (E_{x \sim P}[V(x)] - V(B)) | \\
    &= |\alpha^\ast (\alpha x + (1-\alpha) y - x) + (1-\alpha^\ast) (\alpha x + (1-\alpha) y - y)| \\
    &= |\alpha^\ast (\alpha-1)(x-y) + (1-\alpha^\ast) \alpha(x-y)| \\
    &= |\alpha - \alpha^\ast| |x-y|
\end{align*}
Clearly $\min_{\alpha \in [0,1]} \max_{x,y \in [0,M]} |\alpha - \alpha^\ast| |x-y| = 0$ where $\alpha = \alpha^\ast$.
\end{proof}
\end{exmp}

We do not have to worry about the choice of norm for MML because we know that the OPE error is precisely $\Loss_{MML}$. On the other hand, as shown in the example, this is not the case for VAML.

\newpage
\section{OPO}


In this section we explore the OPO results in the order in which they were presented in the main paper. 

\subsection{Main Result}
\label{app:opo:sec:main}

\begin{proof}
[Proof for Theorem \ref{thm:learningexact}] Fix some $P \in \M$. Through addition of $0$, we get
\begin{align*}
    J(\pi^\ast_{P^\ast}, P^\ast) - J(\pi^\ast_{P}, P^\ast) &=
    J(\pi^\ast_{P^\ast}, P^\ast) - J(\pi^\ast_{P^\ast}, P)  \\
    &\quad+  J(\pi^\ast_{P^\ast}, P) - J(\pi^\ast_{P}, P)  \\ 
    &\quad\quad+ J(\pi^\ast_{P}, P) - J(\pi^\ast_{P}, P^\ast)
\end{align*}
Since $\pi_{P}^\ast$ is optimal in $P$ then $J(\pi^\ast_{P^\ast}, P) - J(\pi^\ast_{P}, P) \leq 0$ which implies
\begin{align*}
    J(\pi^\ast_{P^\ast}, P^\ast) - J(\pi^\ast_{P}, P^\ast) &\leq
    J(\pi^\ast_{P^\ast}, P^\ast) - J(\pi^\ast_{P^\ast}, P) + J(\pi^\ast_{P}, P) - J(\pi^\ast_{P}, P^\ast)
\end{align*}
Taking the absolute value of both sides, triangle inequality and invoking Lemma \ref{thm:opeerror} yields:
\begin{align*}
    |J(\pi^\ast_{P^\ast}, P^\ast) - J(\pi^\ast_{\widehat{P}}, P^\ast)| &\leq 2 \gamma \max_{w,V} | L(w, V, \widehat{P}) | = 2 \gamma \min_{P} \max_{w,V} | L(w,V,P) |
\end{align*}
when $w^{P^\ast}_{\pi^\ast_{P^\ast}},w^{P^\ast}_{\pi^\ast_{P}} \in \mathcal{W}$ and $V^{P}_{\pi^\ast_{P^\ast}}, V^{P}_{\pi^\ast_{P}} \in V$ for every $P \in \M$, or alternatively $w^{P}_{\pi^\ast_{P^\ast}},w^{P}_{\pi^\ast_{P}} \in \mathcal{W}$ and $V^{P^\ast}_{\pi^\ast_{P^\ast}}, V^{P^\ast}_{\pi^\ast_{P}} \in V$ for every $P \in \M$.
\end{proof}

\subsection{Sample Complexity for OPO}\label{app:opo:sec:samplecomplexity}

Since we will only have access to the empirical version $\widehat{P}_n$ rather than $\widehat{P}$, we provide the following bound

\begin{thm}[Learning Error] \label{thm:learning}
Let the functions in $\V$ and $\W$ be uniformly bounded by $C_V$ and $C_W$ respectively. Assume the conditions of Theorem \ref{thm:learningexact} hold and $|\mathcal{R}| \leq \Rmax, \gamma \in [0,1)$. Then, with probability $1-\delta$, 
\begin{align*}
|J(\pi^\ast_{\widehat{P}_n}, &P^\ast) - J(\pi^\ast_{P^\ast}, P^\ast)| \leq 2 \gamma \min_{P}\max_{w,V} |L(w, V, P)| \\
&+  8 \gamma \mathfrak{R}_n(\W,\V,\M) + 8 \gamma C_\W C_\V \sqrt{\log(2/\delta)/n}
\end{align*}
where $\mathfrak{R}_n(\W, \V, \M)$ is the Rademacher complexity of the function class 
\begin{align*}
\{(s,a,s') \mapsto   &w(s,a)(E_{x\sim P}[V(x)] - V(s')) : \\ &w \in \mathcal{W}, P \in \mathcal{\M}, V \in \mathcal{V}\}.
\end{align*}
\end{thm}


\begin{proof}
[Proof for Theorem \ref{thm:learning}]
By Theorem \ref{thm:learningexact}, 
$$ |J(\pi^\ast_{\widehat{P}_n }, P^\ast) - J(\pi^\ast_{P^\ast}, P^\ast)| \leq 2 \gamma \max_{w,V} | L(w, V, \widehat{P}_n) | .$$
We have shown in the proof of Theorem \ref{thm:opeerror} that
$$ \max_{w,V}| L(w,V,\widehat{P}_n) | \leq \min_{P}\max_{w,V} |L(w, V, P)| +  4 \mathfrak{R}_n(\W, \V, \M) + 4  C_\W C_\V \sqrt{\log(2/\delta)/n}. $$
Combining the two completes the proof.
\end{proof}

This bound has the same interpretation as in the OPO case, see Section \ref{app:ope:sec:samplecomplexity}.

\newpage
\subsection{Misspecification}
\label{app:opo:sec:misspec}

Similarly as in Section \ref{app:ope:sec:misspec}, we show the misspecification gap for OPO in the following result.

\begin{lem}[OPO Misspecification]\label{app:lem:misspecOPO}
Let $\Hh \subset (\State \times \A \times \State \to \R)$ be functions on $(s,a,s')$. Denote $(WV)^{\ast}_{P^\ast} = w^{P^\ast}_{\pi^\ast_{P^\ast}}(s,a) V^P_{\pi^\ast_{P^\ast}}(s')$ and $(WV)^{\ast}_{P} = w^{P^\ast}_{\pi^\ast_{P}}(s,a) V^P_{\pi^\ast_{P}}(s')$.
\begin{equation}
    |J(\pi, \widehat{P}) - J(\pi, P^\ast)| \leq 2 \gamma \left(\min_{P \in \M} \max_{h \in \Hh} |\Loss(h,P)| + \epsilon_\Hh \right)
\end{equation}
where $\epsilon_\Hh = \max(\max_{P \in \M} \min_{h\in \Hh} |\Loss((WV)^\ast_{P^\ast} - h, P)|, \max_{P \in \M} \min_{g \in \Hh} |\Loss((WV)^\ast_{P} - g, P)|).$

\end{lem}
\begin{proof}
[Proof for Lemma \ref{app:lem:misspecOPO}]
From the proof of Theorem \ref{thm:learningexact}, $J(\pi^\ast_{P^\ast}, P^\ast) - J(\pi^\ast_{P}, P^\ast) \leq J(\pi^\ast_{P^\ast}, P^\ast) - J(\pi^\ast_{P^\ast}, P) + J(\pi^\ast_{P}, P) - J(\pi^\ast_{P}, P^\ast) = \Loss(w^{P^\ast}_{\pi^\ast_{P^\ast}}, V^P_{\pi^\ast_{P^\ast}},P) + \Loss(w^{P^\ast}_{\pi^\ast_{P}}, V^P_{\pi^\ast_{P}},P)$. 
Using the result from proof of Lemma \ref{lem:misspec},  
\begin{align*}
    |\Loss(w^{P^\ast}_{\pi^\ast_{P^\ast}}, V^P_{\pi^\ast_{P^\ast}},P) + \Loss(w^{P^\ast}_{\pi^\ast_{P}}, V^P_{\pi^\ast_{P}},P)| &\leq |\Loss(h,P) + \Loss((WV)^{\ast}_{P^\ast}-h,P)| + |\Loss(g,P) + \Loss((WV)^{\ast}_{P}-g,P)| \\
    &\leq 2 \min_P \max_{h \in \Hh} |\Loss(h,P)| + \max_P \min_{h \in \Hh} |\Loss((WV)^\ast_{P^\ast} - h, P)| \\
    & \quad\quad\quad\quad\quad\quad\quad\quad\quad\quad\quad\quad\quad\quad + \max_P \min_{g \in \Hh} |\Loss((WV)^\ast_{P} - g, P)| \\
    &\leq 2 (\min_P \max_{h\in \Hh} |\Loss(h,P)| + \epsilon_\Hh)
\end{align*}
where $\epsilon_\Hh = \max(\max_P \min_h |\Loss((WV)^\ast_{P^\ast} - h, P)|, \max_P \min_g |\Loss((WV)^\ast_{P} - g, P)|).$
Therefore $|J(\pi, \widehat{P}) - J(\pi, P^\ast)| \leq 2 \gamma (\min_P \max_h |\Loss(h,P)| + \epsilon_\Hh),$ as desired.
\end{proof}
\newpage
\section{Additional theory}
\label{app:ope/opo:sec:additional}

In this section, we provide additional results that were not covered in the paper. Specifically, we show that as we make $\W,\V$ too rich then the only model with zero loss is $P^\ast$ itself, which may not be in $\M$.

\subsection{Necessary and sufficient conditions for uniqueness of $|\Loss(w,V,P)| = 0$}

When $\W,\V$ are in $L^2$ then $|\Loss| = 0$ is uniquely determined: 
\begin{lem}[Necessary and Sufficient]\label{lem:nec&suff} $\Loss(w,V,P) = 0$ for all  $w \in L^2(\X, \nu) = \{g: \int g^2(x,a) d\nu(x,a) < \infty \}, V \in L^2(\State, \nu) = \{f: \int f^2(x) d\nu(x) < \infty \}$ if and only if $P = P^\ast$ wherever $D_{\pi_b}(s,a) \neq 0$. 
\end{lem}

\begin{cor}\label{cor:nec&suff}The same result holds if $w\cdot V \in L^2(\X \times \State, \nu) = \{h: \int h^2(x,a,x') d\nu(x,a,x') < \infty \}$.
\end{cor}


\begin{proof}[Proof for Lemma \ref{lem:nec&suff} and Corollary \ref{cor:nec&suff}] 
We begin with definition \ref{loss:mml} and expand the expectation.
\begin{align*}
L(w,V,P) = &E_{(s,a,s') \sim D_{\pi_b}(\cdot,\cdot) P^\ast(\cdot|s,a)} [w(s,a) \left( E_{x\sim P(\cdot|s,a)}[V(x)] - V(s') \right)] \\
&=E_{(s,a) \sim D_{\pi_b}(\cdot,\cdot)} [w(s,a) \left( E_{s'\sim P(\cdot|s,a)}[V(s')] - E_{s' \sim P(\cdot|s,a)}[V(s')] \right)] \\
&= \int D_{\pi_b}(s,a)  w(s,a)(V(s') (P(s'|s,a)  - P^\ast(s'|s,a)) \; d\nu(s,a,s').
\end{align*}
($\Rightarrow$) Clearly if $P = P^\ast$ then $L(w,V,P) = 0$.
($\Leftarrow$) For the other direction, suppose $L(w,V,P) = 0$. By assumption, $w(s,a)$ can take on any function in $L^2(\X, \nu)$ and therefore if $L(w,V,P) = 0$ then 
\begin{equation}\label{eq:VP-P} \int V(s') (P(s'|s,a)  - P^\ast(s'|s,a)) \; d\nu(s') = 0,
\end{equation}
wherever $D_{\pi_b}(s,a) \neq 0$.
Similarly, $V(s')$ can take on any function in $L^2(\State, \nu)$ and therefore if equation (\ref{eq:VP-P}) holds then $P = P^\ast$. 
For the corollary, let $(w,V) \in \W\V$ take on any function in $L^2(\X \times \State, \nu)$. If $L(w,V,P) = 0$ then 
$P(s'|s,a)  - P^\ast(s'|s,a) = 0$, as desired.
\end{proof}

In an RKHS, when the kernel corresponds to an integrally strict positive definite kernel (ISPD), $P=P^\ast$ remains the unique minimizer of the MML Loss:

\begin{lem}[Realizability means zero loss even in RKHS]\label{lem:rkhsnec&suff}
$\Loss(w,f,P) = 0$ if and only if $P = P^\ast$ for all $(w,V) \in \{(w(s,a), V(s')) : \langle w V, wV \rangle_{\mathcal{H}_k} \leq 1, w: X \times A \to \mathbb{R}, V:  X \to \mathbb{R} \}$ in an RKHS with an integrally strict positive definite (ISPD) kernel.
\end{lem}


\begin{proof}
[Proof for Lemma \ref{lem:rkhsnec&suff}] 

\cite{uehara2019minimax} prove an analogous result and proof here is included for reader convenience. From Mercer’s theorem \cite{mohri2012foundations}, there exists an orthonormal basis $(\phi_j)_{j=1}^\infty$ of $L^2(\X \times \State, \nu)$ such that RKHS is represented as
$$ \W\V = \left\{w\cdot V = \sum_{j=1}^\infty b_j \phi_j \;\bigg| (b_j)_{j=1}^\infty \in l^2(\mathbb{N}) \mbox{ with } \sum_{j=1}^\infty \frac{b_j^2}{\mu_j} < \infty  \right\} $$
where each $\mu_j$ is a positive value since kernel is ISPD. Suppose there exists some $P \in \M$ such that $L(w,V,P)=0$ for all $(w,V) \in \W\V$ and $P \neq P^\ast$. Then, by taking $b_j = 1$ when $(j = j')$ and  $b_j = 0$ when $(j \neq j')$ for any $j' \in \mathbb{N}$, we have $L(\phi_j, P)= 0$ where we treat $w \cdot V$ as a single input to $L$. This implies $L(w, V, P) = 0$ for all $w\cdot V \in L^2(\X \times \State , \nu) = 0$. This contradicts corollary \ref{cor:nec&suff}, concluding the proof.
\end{proof}
\newpage

\section{Scenarios \& Considerations}
\label{app:sec:scenarios}

In this section we give proof for the various propositions for the corresponding section in the main paper.

\subsection{Linear Function Classes}
\label{app:scenarios:linfunc}

\begin{proof}
[Proof for Prop. \ref{lem:linearfunc}]
Given $w(s,a)V(s') = \psi(s,a,s')^T \beta$ and $P(s'|s,a) = \phi(s,a,s')^T \alpha$ then
\begin{align*}
    L_n(w,V,P) &= 
    E_{n} [ E_{x \sim P}[ \psi(s,a,x)^T \beta] - \psi(s,a,s')^T \beta], \\
    &= E_{n} \left[ \int \alpha \phi(s,a,x)^T  \psi(s,a,x)^T \beta d\nu(x) - \psi(s,a,s')^T \beta \right], \\
    &= E_{n} [ \alpha^T \left( \int \phi(s,a,s') \psi(s,a,s')^T d\nu(s') \right) \beta - \psi(s,a,s')^T \beta], 
\end{align*}
which is linear in $\beta$. $L_n^2(w,V,P) = 0$ is achieved through $E_{n} [ \alpha^T \left( \int \phi(s,a,s') \psi(s,a,s')^T d\nu(s') \right) - \psi(s,a,s')^T] = 0$. Thus, 
$$
\widehat{\alpha}^T = E_n[\psi(s,a,s')^T] E_n\left[\int \phi(s,a,s') \psi(s,a,s')^T d\nu(s') \right]^{-1},
$$ assuming $E_n\left[\int \phi(s,a,s') \psi(s,a,s')^T d\nu(s') \right]$ is full rank.
Taking the transpose completes the proof.
\end{proof}


\begin{proof}
[Proof for Prop. \ref{lem:tabular}]
We begin with $\phi(s,a,s') = e_{(s,a,s')}$, the (s,a,s')-th standard basis vector and $\psi = \phi$. Then 
$$ X(s,a) = (\sum_{x \in \State} \phi(s,a,x) \phi(s,a,x)^T)_{i,j} =  \begin{cases} 
      1 & i = s|\A||\State| + a|\State|, i=j \\
      0 & \text{otherwise}
   \end{cases}
.$$ Notice that $X(s,a)$ is a diagonal matrix and is the discrete counter-part to  $\int \phi(s,a,s') \psi(s,a,s')^T d\nu(x)$. Therefore, $E_n[X(s,a)] = \frac{1}{N} \sum_{(s,a,s') \in D} X(s,a),$ which is
a diagonal matrix of the average number of times $(s,a)$ appears in the dataset $D$. Similarly, $E_n[\phi(s,a,s')]$ is the average number of times that $(s,a,s')$ appears in the dataset $D$. Hence, by Prop \ref{lem:linearfunc}, 
$$ \widehat{\alpha}_{s,a,s'} = \frac{\#\{(s,a,s') \in D\}}{\#\{(s,a,x) \in D: \forall x \in \State \}}.
$$
Therefore $P(s'|s,a) = \phi(s,a,s')^T \widehat{\alpha} = \widehat{\alpha}_{s,a,s'},$ as desired.

\end{proof}


\subsection{LQR}
\label{app:scenarios:lqr}

In order to provide proof that MML gives the LQR-optimal solution, we begin with a few Lemmas. First, we show that the value function is quadratic.

\begin{lem}[Value Function is Quadratic]\label{lem:VQuad_LQR} Let $s_{t+1} = As_t + Ba_t + w$ with $w \sim N(0, \sigma^{\ast 2} I)$ be the dynamics, $\pi_K(a|s) = -Ks + w_K$ where $w_K \sim N(0, \sigma_K^2 I)$ be the policy. Let $\gamma \in (0,1]$ be the discount factor. Then $V(s) = s^T U s + q$ where
\begin{align*}
    U &= Q + K^TRK + \gamma (A-BK)^T U (A-BK) \\
    q &= \frac{1}{1-\gamma}(\sigma^2_K tr(R) + \gamma \sigma_K^2 tr(B^T U B) + \gamma \sigma^{\ast 2} tr(U)).
\end{align*}
\end{lem}

\begin{proof}
[Proof for Lemma \ref{lem:VQuad_LQR}]
The value function is given by:
\begin{align*}
    x^T U x + q &= x^T Q x + E_{N(-K x, \sigma_K^2 I)}[u^T R u + \gamma E_{N(A x + B u, \sigma^{\ast2} I)}[V(s')]] \\
    &= x^T Q x + E_{N(-K x, \sigma_K^2 I)}[u^T R u + \gamma (Ax + Bu)^T U (Ax + Bu) + \gamma q + \gamma \sigma^{\ast 2} tr(U) ] \\
    &= x^T Q x + x^T K^T R K x + \sigma^2_K tr(R) + \gamma x^T(A-BK)^T J (A-BK)x \\
    &\quad\quad\quad\quad + \gamma \sigma_K^2 tr(B^T U B) + \gamma q + \gamma \sigma^{\ast2} tr(U)
\end{align*}
Thus, the quadratic terms satisfy
$$ U = Q + K^TRK + \gamma (A-BK)^T U (A-BK) $$
and the linear term satisfies
$$ q = \frac{1}{1-\gamma}(\sigma^2_K tr(R) + \gamma \sigma_K^2 tr(B^T U B) + \gamma \sigma^{\ast2} tr(U))
$$
The final value is given by:
$$ J(\pi, P^\ast) = E_{N(s_0,\sigma_0^2 I)}[U] = s_0^T U s_0 + q+ \sigma_0^2 tr(U)$$
Existence and uniqueness of $U,q$ is heavily studied \citep[]{bertsekas2005dynamic}.
\end{proof}


Under the same assumptions as Lemma \ref{lem:VQuad_LQR}, we can simplify $\Loss$ into a reduced form:

\begin{lem}[LQR Loss Simplified]\label{lem:L_LQR} In addition to the assumptions of Lemma \ref{lem:VQuad_LQR}, let $d_0 = s_0 + w_{d_0}$ where $w_{d_0} \sim N(0, \sigma_{d_0}^2 I)$ be the initial state distribution. Let $P = A s + B a \in \M$ where $A \in \R^{n \times n}, B \in \R^{n \times k}$ 
and $(A,B)$ is controllable. Let $K \in \R^{k \times n}$ represent all linear policies and $U \in \mathbb{S}_{+}^{n}$ be all symmetric positive semi-definite matrices.
\begin{align*}
    &\min_P \max_{w,V} |\Loss(w,V,P)| \\ &= \min_{A,B} \max_{K, U} \sum_i \gamma^i [s_0^T(A^\ast - B^\ast K)^{i T}\Delta(A^\ast - B^\ast K)^i s_0 \\
    &\quad\ + tr(\Delta \Sigma_i)] + \sigma_K^2 tr(B^TUB - B^{\ast T}UB^\ast) - \sigma^{\ast  2} tr(U),
\end{align*}
where $\Delta = (A-BK)^{T}U(A-BK) - (A^\ast-B^\ast K)^{ T}U(A^\ast-B^\ast K)$ and $\Sigma_i = \sigma^\ast (I + \ldots + F^{i-1}F^{(i-1)T}) + \sigma_K (B^\ast B^{\ast T} + \ldots + F^{i-1}B^\ast B^{\ast T}F^{(i-1)T}) + \sigma_0 F^{i} F^{iT} $ for $i > 0$ and $\Sigma_0 = \sigma_0 I$, $F = A^\ast - B^\ast K$.
\end{lem}

\begin{proof}
[Proof for Lemma \ref{lem:L_LQR}]
We first show that the evolution of dynamics $P^\ast$ under gaussian noise, with a linear gaussian controller is a gaussian mixture $\sum_{i} N((A^\ast - B^\ast K)^i s_0, \Sigma_i)$, where $\Sigma_i = \sigma^\ast (I + \ldots + F^{i-1}F^{(i-1)T}) + \sigma_K (B^\ast B^{\ast T} + \ldots F^{i-1}B^\ast B^{\ast T}F^{(i-1)T}) + \sigma_0 F^{i} F^{iT} $ for $i > 0$ and $\Sigma_0 = \sigma_0 I$, $F = A^\ast - B^\ast K$. 

It's clear $s_0 \sim N(s_0, \sigma_0^2 I)$, the base case. Suppose for induction $s_n \sim N((A^\ast - B^\ast K)^n s_0, \Sigma_n)$ holds for some $n \geq 0$. Then
\begin{align*}
    s_{n+1} &= A^\ast s_n + B^\ast (-K s_n + w_K) + w^\ast \\
    &= (A^\ast - B^\ast K) s_n + B^\ast w_k + w^\ast \\
    &\sim N((A^\ast - B^\ast K)^{n+1} s_0, (A^\ast - B^\ast K)\Sigma_n(A^\ast - B^\ast K)^T + B^\ast B^{\ast T} + \sigma^\ast I) \\
    &= N((A^\ast - B^\ast K)^{n+1} s_0, \Sigma_{n+1}),
\end{align*}
completing the inductive step. Notice every step $s_t$ is sampled from a gaussian distribution, therefore \begin{equation}\label{eq:lqr_dpi}
    d_{\pi,\gamma}^{P^\ast}(s,a) = \sum_{i=0}^\infty \gamma^i N(s; F^i s_0, \Sigma_i) N(a; -K s, \sigma_K^2 I),
\end{equation} is a gaussian mixture.
Let $w = \frac{d_{\pi,\gamma}^{P^\ast}}{D}$. We know $V$ is quadratic, given by $U \in \mathcal{S}_+^n.$ Therefore,
\begin{align*}
&\min_{P} \max_{w,V} \Loss(w, V, P) = \min_{A,B} \max_{w, V}E_{(s,a)\sim D}[w [E_P[V] - E_{P^\ast}[V] ]]\\  
    &=\min_{A,B} \max_{w, U}E_{(s,u)\sim D}[w [ (As+Bu)^T U (As+Bu) - (A^\ast s + B^\ast u)^T U (A^\ast s + B^\ast u) - \sigma^{\ast  2} tr(U)]] \\
    &=\min_{A,B} \max_{K, U} E_{\sum_i \gamma^i N((A^\ast - B^\ast K)^i s_0, \Sigma_i)}[E_{u\sim N(-Ks,\sigma^2_k I)}[ w [ (As+Bu)^T U (As+Bu) \\
    &\quad\quad\quad\quad\quad\quad\quad\quad\quad\quad\quad\quad\quad\quad\quad\quad\quad\quad\quad\quad\quad - (A^\ast s + B^\ast u)^T U (A^\ast s + B^\ast u) - \sigma^{\ast  2} tr(U)] ]] \\
    &=\min_{A,B} \max_{K, U} E_{\sum_i \gamma^i N((A^\ast - B^\ast K)^i s_0, \Sigma_i)}[s^T[(A-BK)^{T}U(A-BK) - (A^\ast-B^\ast K)^{ T}U(A^\ast-B^\ast K)] s \\
    &\quad\quad\quad\quad\quad\quad\quad\quad\quad\quad\quad\quad\quad\quad\quad\quad\quad\quad\quad\quad\quad + \sigma_K^2 tr(B^TUB) - \sigma_K^2 tr(B^{\ast T}UB^\ast) - \sigma^{\ast  2} tr(U)] \\
    &=\min_{A,B} \max_{K, U} E_{\sum_i \gamma^i N((A^\ast - B^\ast K)^i s_0, \Sigma_i)}[s^T[\Delta(A,B,A^\ast,B^\ast,U,K)] s  + \sigma_K^2 tr(B^TUB - B^{\ast T}UB^\ast) - \sigma^{\ast  2} tr(U)] \\
    &= \min_{A,B} \max_{K, U} \sum_i \gamma^i [s_0^T(A^\ast - B^\ast K)^{i T}\Delta(A^\ast - B^\ast K)^i s_0 + tr(\Delta \Sigma_i)] + \sigma_K^2 tr(B^TU B - B^{\ast T}U B^\ast) - \sigma^{\ast  2} tr(U)
\end{align*}
where $\Delta = (A-BK)^{T}U(A-BK) - (A^\ast-B^\ast K)^{ T}U(A^\ast-B^\ast K).$
\end{proof}

First, Lemma \ref{lem:L_LQR} supposes that there is model mismatch $P^\ast \not\in \M$ since $\M$ are deterministic simulators and $P^\ast$ is stochastic. Second, we notice that $K$ takes the position of $w$, which is to say that the policy $K$ directly specifies $w$, as expected. We will need the previous two results in the experiments. We may now prove Prop \ref{lem:MML_MLE_LQR} that says MML yields the true parameters of LQR in expectation: 


\begin{proof}
[Proof for Prop \ref{lem:MML_MLE_LQR}]

Consider two linear, controllable systems with parameters $P_1 = (A_1, B_1)$ and $P_2 = (A_2, B_2)$. Then there exists a controller $K$ that stabilizes $P_1$ (i.e, $J(P_1), K) < \infty$) but destabilizes $P_2$ (i.e, $J(P_2, K) = \infty$). We show this by analyzing the characteristic polynomial of both $A_1-B_1 K$ and $A_2 - B_2 K$. There exists an invertible matrix $T_1, T_2$ that put $(A_1, B_1), (A_2,B_2)$ into controllable canonical forms (CCF), respectively \cite{bertsekas2005dynamic}. Thus, we will assume, wlog, that $(\tilde{A}_1, \tilde{B}_1), (\tilde{A}_2, \tilde{B}_2)$ are already in CCF. Hence,
$$
\tilde{A}_1 = \begin{bmatrix}
0 & 1 & 0 & \ldots & 0 \\
0 & 0 & 1 & \ldots & 0 \\
\vdots & \vdots & \vdots &  & \vdots \\
0 & 0 & 0 &  & 1 \\
-a_0 & -a_1 & -a_2 & \ldots & -a_{n-1} \\
\end{bmatrix},
\quad 
\tilde{B}_1 = \begin{bmatrix}
0  \\
0  \\
\vdots \\
0 \\
1 \\
\end{bmatrix}
$$
and 
$$
\tilde{A}_2 = \begin{bmatrix}
0 & 1 & 0 & \ldots & 0 \\
0 & 0 & 1 & \ldots & 0 \\
\vdots & \vdots & \vdots &  & \vdots \\
0 & 0 & 0 &  & 1 \\
-b_0 & -b_1 & -b_2 & \ldots & -b_{n-1} \\
\end{bmatrix},
\quad 
\tilde{B}_2 = \begin{bmatrix}
0  \\
0  \\
\vdots \\
0 \\
1 \\
\end{bmatrix}
$$
We will find a controller in the form $K = K_1 T_1 = K_2 T_2$ for some $K_1,K_2$ for $T_1,T_2$ that put the systems into CCF. Consider a desired characteristic polynomial of $f(s) = (s+\epsilon)^{n-1} (s+\lambda)$ for $\epsilon, \lambda \in \R^+ (> 0)$. This polynomial has eigenvalues equal to $-\epsilon, -\lambda$ and therefore a system with this polynomial is asymptotically stable (converges to $0$ exponentially fast). Take $K_1 = [k_{1,0}, k_{1,1},\ldots,k_{1,n-1}]$. Then $det(sI - (\tilde{A}_1 - \tilde{B}_1 K_1)) = s^n + (a_{n-1} + k_{1,n-1})s^{n-1} + \cdots + (a_0 + k_{1,0}).$ By selecting $k_{1,i} = \left( {n-1 \choose i} \lambda + {n-1 \choose i-1} \epsilon \right) \epsilon^{n-1-i} - a_i$ then $det(sI - (\tilde{A}_1 - \tilde{B}_1 K_1)) = f(s)$. Hence, $(\tilde{A}_1, \tilde{B}_1)$ is asymototically stable with eigenvalues $-\lambda, -\epsilon$ for any $\lambda,\epsilon$ strictly positive. Therefore $K = K_1 T_1$ makes the system $(A_1, B_1)$ asymptotically stable. 

Now we consider $K_2 = K_1 T_1 T_2^{-1}$. Let us denote $T_1 T_2^{-1} = T$ which is also invertible since $T_1,T_2$ are invertible. Then by taking the last term of $det(sI - (\tilde{A}_2 - \tilde{B}_2 K_2))$, we can examine the product $\prod_{i=0}^{n-1} \lambda_i$ of the eigenvalues of the closed loop system $\tilde{A}_2 - \tilde{B}_2 K_2$. Namely, $b_0 + \sum_{i=0}^{n-1} k_{1,i} T_{i,n}$ is the product of eigenvalues. We may simplify this via some algebra as follows:
\begin{align*}
    \prod_{i=0}^{n-1} \lambda_i &= b_0 + \sum_{i=0}^{n-1} k_{1,i} T_{i,n} \\
    &= b_0 + \sum_{i=0}^{n-1} T_{i,n} \left(\left( {n-1 \choose i} \lambda + {n-1 \choose i-1} \epsilon \right) \epsilon^{n-1-i} - a_i\right) \\
    &= \underbrace{b_0 - \sum_{i=0}^{n-1} a_i  + \sum_{i=0}^{n-1} T_{i,n} {n-1 \choose i-1} \epsilon^{n-i}}_{\bar{b}} + \lambda \underbrace{\sum_{i=0}^{n-1} T_{i,n} {n-1 \choose i} \epsilon^{n-1-i}}_{c} \\
    &= \bar{b} + \lambda c
\end{align*}

We may select $\epsilon > 0$ so that $c \neq 0$ otherwise $T_{i,n} = 0$ for all $i$ which would contradict invertibility of $T$. Therefore $\prod_{i=0}^{n-1} \lambda_i$ is linear in $\lambda$. By driving $\lambda \to \infty$,  then $|\prod_{i=0}^{n-1} \lambda_i| \to \infty$ is unbounded. Select $\lambda$ so that $|\bar{b} + \lambda c| > 1$. By the pigeonhole principle, at least one of the eigenvalues of $\tilde{A}_2 - \tilde{B}_2 K_2$ must have a magnitude greater than $1$ and therefore the system is unstable. Therefore the controller $K_2 T_2 = K_1 T_1 T_2^{-1} T_2 = K_1 T_1 = K$ makes the system $(A_2,B_2)$ unstable. Hence, $K$ simultaneously stabilizes $(A_1,B_1)$ but destabilizes $(A_2,B_2)$.


According to Lemma \ref{lem:L_LQR}, when $(A,B) = (A^\ast, B^\ast)$ then for any $K$, $\max_{U} \Loss((A,B), K, U) = \max_{U} |\sigma^{\ast 2} tr(U)| < \infty$  since $U$ are bounded by assumption. Furthermore, we have just shown that there always exists a $K$ that destabilizes any controller $(A,B) \neq (A^\ast, B^\ast)$ while stabilizing $(A^\ast, B^\ast)$. Therefore $\max_{K,U} \Loss((A,B), K, U) = \infty$ for any system $(A,B) \neq (A^\ast, B^\ast).$ Therefore $\min_{(A,B)} \max_{K,U} \Loss((A, B), K, U) = (A^\ast, B^\ast)$. 

It is well known that ordinary least squares is a consistent estimator when the noise is exogenous, as it is here. Therefore the maximum likelihood solution also yields $(A^\ast, B^\ast)$ in expectation.
\end{proof}



\subsection{RKHS \& Practical Implementation}
\label{app:scenarios:rkhs}

Since $P \in \M$ is a stochastic model in general, then the inner expectation of the loss in def (\ref{loss:mml}) over $P$ involves sampling $x$ from $P(\cdot|s,a)$ and computing the empirical average of $V(x)$. In general this can be computationally demanding if $\State$ is high dimensional and $P$ does not have a closed form, requiring MCMC estimates or variational estimates \citep[]{MacKay2002MCMC,Goodfellow2016DL}. However, in practice, most parametrizations of models use nice distributions, such as gaussians, from which sampling is efficient. This issue is similarly present in other decision-aware literature  \citep[e.g.,][]{farahmand2017VAML}. 

The estimator based on Eq (\ref{phat}) requires solving a minimax problem which is often computationally challenging. One approach might be to set-up neural networks in a GAN-like fashion and use a higher order gradient descent \citep[]{ Goodfellow2014GAN, Schaefer2019CGD}. 

If we have access to a kernel, say radial basis function (RBF), then the inner maximization over $w,V$ has a closed form when $\W \times \V$ correspond to a reproducing kernel Hilbert space (RKHS), $H_K$ with kernel $K$. In particular, in similar spirit to \citep[]{liu2018breaking, feng2019kernel, uehara2019minimax} we have

\begin{prop}[Closed form exists in RKHS]\label{lem:rkhs} Assume $\mathcal{WV} = \{(w(s,a), V(s')) : \langle w V, wV \rangle_{\mathcal{H}_K} \leq 1, w: \X \to \mathbb{R}, V:  \State \to \mathbb{R} \}$. Let $\langle \cdot, \cdot \rangle_{\mathcal{H}_K}$ be an inner product on $\mathcal{H}_K$ satisfying the reproducing kernel property $w(s,a)V(s') = \langle w V, K((s,a,s'), \cdot) \rangle_{\mathcal{H}_K}.$ The term $\max_{(w,V) \in \mathcal{WV}} \Loss(w,V,P)^2$ has a closed form:
\begin{align*}
\max_{(w,V) \in \mathcal{WV}} &\Loss(w,V,P)^2 = E_{(s,a,s')\sim D_{\pi_b} P^\ast, (\tilde{s},\tilde{a},\tilde{s}')\sim D_{\pi_b} P^\ast}\bigg[\\
&E_{x\sim P, \tilde{x}\sim P}[ K((s,a,x),(\tilde{s}, \tilde{a}, \tilde{x}))]\\
& - 2
E_{x\sim P}[ K((s,a,x),(\tilde{s}, \tilde{a}, \tilde{s}'))] \\
&+ K((s,a,s'),(\tilde{s}, \tilde{a}, \tilde{s}')) \bigg]
\end{align*}
\end{prop}
\begin{proof}[Proof for Prop \ref{lem:rkhs}] 
Recall that by the reproducing property of kernel $K$ in the RKHS space $H_K$ then $\langle f, K \rangle_{H_K}$ for any $f \in H_K$. Starting from definition \ref{loss:mml},
\begin{align*}
L(w,V,P)^2 = &E_{(s,a,s') \sim D_{\pi_b}(\cdot,\cdot) P^\ast(\cdot|s,a)} [w(s,a) \left( E_{x\sim P(\cdot|s,a)}[V(x)] - V(s') \right)]^2 \\
&=E_{(s,a,s',x) \sim D_{\pi_b}(\cdot,\cdot) P^\ast(\cdot|s,a)P(\cdot|s,a)} [w(s,a) V(x) - w(s,a) V(s') ]^2 \\
&= E_{(s,a,s',x) \sim D_{\pi_b}(\cdot,\cdot) P^\ast(\cdot|s,a)P(\cdot|s,a)} [\langle w V, K((s,a,x), \cdot) \rangle_{\mathcal{H}_k} -  \langle w V, K((s,a,s'), \cdot) \rangle_{\mathcal{H}_k}]^2 \\
&= \langle wV, (wV)^\ast \rangle_{\mathcal{H}_k}^2
\end{align*}
where $(wV)^\ast(\cdot) =  E_{(s,a,s',x) \sim D_{\pi_b}(\cdot,\cdot) P^\ast(\cdot|s,a)P(\cdot|s,a)} [K((s,a,x), \cdot) - K((s,a,s', \cdot)].$ By Cauchy-Schwarz and the fact that $wV$ is within a unit ball,  then
$$ \max_{w,V \in \mathcal{WV}} L(w,f,V)^2 = \max_{w,V \in \mathcal{WV}} \langle wV, (wV)^\ast \rangle_{\mathcal{H}_k}^2 = \| (wV)^\ast \|^2 = \langle (wV)^\ast, (wV)^\ast \rangle_{\mathcal{H}_k}. $$
Expanding,
\begin{align*}
    \max_{w,V \in \mathcal{WV}} L(w,f,V)^2 &= \langle (wV)^\ast, (wV)^\ast \rangle_{\mathcal{H}_k} \\
    &= \langle E_{(s,a,s',x) \sim D_{\pi_b}(\cdot,\cdot) P^\ast(\cdot|s,a)P(\cdot|s,a)} [K((s,a,x), \cdot) - K((s,a,s', \cdot)], \\
    &\quad\quad E_{(\tilde{s},\tilde{a},\tilde{s}',\tilde{x}) \sim D_{\pi_b}(\cdot,\cdot) P^\ast(\cdot|\tilde{s},\tilde{a})P(\cdot|\tilde{s},\tilde{a})} [K((\tilde{s},\tilde{a},\tilde{x}), \cdot) - K((\tilde{s},\tilde{a},\tilde{s}', \cdot)] \rangle_{\mathcal{H}_k} \\
    &= \bigg\langle \int D_{\pi_b}(s,a) P^\ast(s'|s,a) P(x|s,a) (K((s,a,x), \cdot) - K((s,a,s'), \cdot)) , \\
    &\quad\quad \int D_{\pi_b}(\tilde{s},\tilde{a}) P^\ast(\tilde{s}'|\tilde{s},\tilde{a}) P(\tilde{x}|\tilde{s},\tilde{a}) (K((\tilde{s},\tilde{a},\tilde{x}), \cdot) - K((\tilde{s},\tilde{a},\tilde{s}'), \cdot)) , \bigg\rangle_{\mathcal{H}_k} \\
    &= \int D_{\pi_b}(s,a) P^\ast(s'|s,a) P(x|s,a) D_{\pi_b}(\tilde{s},\tilde{a}) P^\ast(\tilde{s}'|\tilde{s},\tilde{a}) P(\tilde{x}|\tilde{s},\tilde{a}) \\
    &\quad\quad \times \left\langle  K((s,a,x), \cdot) - K((s,a,s'), \cdot), K((\tilde{s},\tilde{a},\tilde{x}), \cdot)  - K((\tilde{s},\tilde{a},\tilde{s}', \cdot)  \right\rangle_{\mathcal{H}_k}  \\
\end{align*}
By linearity of the inner product, the reproducing kernel property we get 
\begin{align*}
\max_{(w,V) \in \mathcal{WV}} L(w,f,V)^2 &= E_{(s,a,s',x)\sim D_{\pi_b} P^\ast P, (\tilde{s},\tilde{a},\tilde{s}',\tilde{x})\sim D_{\pi_b} P^\ast P}[
K((s,a,x),(\tilde{s}, \tilde{a}, \tilde{x})) -
K((s,a,x),(\tilde{s}, \tilde{a}, \tilde{s}')) \\
&\quad\quad\quad\quad\quad\quad- K((s, a, s'), (\tilde{s},\tilde{a},\tilde{x}))
+ K((s,a,s'),(\tilde{s}, \tilde{a}, \tilde{s}'))] \\
&= E_{(s,a,s',x)\sim D_{\pi_b} P^\ast P, (\tilde{s},\tilde{a},\tilde{s}',\tilde{x})\sim D_{\pi_b} P^\ast P}[
K((s,a,x),(\tilde{s}, \tilde{a}, \tilde{x})) -
2 K((s,a,x),(\tilde{s}, \tilde{a}, \tilde{s}')) \\
&\quad\quad\quad\quad\quad\quad
+ K((s,a,s'),(\tilde{s}, \tilde{a}, \tilde{s}'))],
\end{align*}
where for the last equality we used the fact that $K$ is symmetric.
\end{proof}
\newpage


\section{Experiments}
\label{app:experiments}

\subsection{Environment Descriptions}

\subsubsection{LQR}

The LQR domain is a 1D stochastic environment with true dynamics: $P^\ast(s'|s,a) = s - .5 a + w^\ast$ where $w^\ast \sim N(0,.01^2)$. We let $x_0 \sim N(1, .1^2)$. The reward function is $R(s,a) = -(s+a)$ and $\gamma = .9$. We use a finite class $\M$ consisting of all deterministic models $\M = \{P_x(s'|s,a) = (1 + x/10) s - (.5 + x/10) a | x \in [0, M]\}$ where we vary $M \in \{2,3,\ldots, 19\}.$  We write $(A^\ast, B^\ast) = P_0(s'|s,a) = A^\ast s + B^\ast a$, the deterministic version of $P^\ast$.

\subsubsection{Cartpole}

We use the standard Cartpole benchmark (OpenAI, \cite{OpenAIgym}). The state space is a tuple $(x,\dot{x}, \theta, \dot{\theta})$ representing the position of the cart, velocity of the cart, angle of the pole and angular velocity of the pole, respectively. The action space is discrete given by pushing the car to the left or pushing the car to the right. We add $N(0,.001^2)$ Gaussian noise to each component of the state to make the dynamics stochastic.  We consider the infinite horizon setting with $\gamma=.98$. The reward function is modified to be a function of angle and location  $R(s,\theta) = (2 - \theta / \theta_{\max}) * (2 - s / s_{\max}) - 1$) rather than 0/1 to make the OPE problem more challenging.

\subsubsection{Inverted Pendulum (IP)}

We consider the infinite horizon setting with $\gamma=.98$. The state space is a tuple $(\theta, \dot{\theta})$ representing the  angle of the pole and angular velocity of the pole, respectively. The action space $\mathcal{A} = \R$ is continuous representing a clockwise or counterclockwise force. The reward function is a clipped quadratic function $R([\theta, \dot{\theta}], a) = \min(((\theta + \pi) \mod 2\pi - \pi)^2 + .1 \dot{\theta}^2 + .001 u^2, 100)$. This IP environment has a Runge-Kutta(4) integrator \citep[]{laypy} rather than Forwrd Euler and, thus, produces more realistic data. The mass of the rod is $.25$ and the length $.5$.








\subsection{Experiment Descriptions}

\subsubsection{LQR OPE/OPO}

\textbf{OPE.} We aim to evaluate $\pi(a|s) = N(1.3 s, .1^2)$. We ensure $V^P_{\pi} \in \V$ for all $P \in \M$ by solving the equations in Lemma \ref{lem:VQuad_LQR}. We ensure $W^{P^\ast}_{\pi} \in \W$ using Equation \eqref{eq:lqr_dpi}.
We derive 1-d equations for VAML analogous to Lemma \ref{lem:L_LQR}). Finally, we know MLE gives $(A^\ast, B^\ast)$ in expectation (see Prop \ref{lem:MML_MLE_LQR}).

\textbf{Metric:} We compute $|(J(\pi,\widehat{P}) - J(\pi,P^\ast))|,$ the OPE error.

\textbf{OPO.} Similarly as in OPE, we ensure that all MML realizability assumptions hold. This means as we increase $\M$ then we have to increase the sizes of both $\W$ and $\V$ now instead of just $\V$ as in OPE.
Once again MLE gives $(A^\ast, B^\ast)$ in expectation (see Prop \ref{lem:MML_MLE_LQR}) and we evaluate VAML using equations analogous to those in Lemma \ref{lem:L_LQR}).
With this, we produce Figure \ref{fig:LQR_MML_vs_rest_OPO} (right). By increasing $\M$, we also have more policies $\{\pi^\ast_{P}\}_{P \in \M}$ we may consider. Instead of selecting one for OPE, for each $\pi \in \{\pi^\ast_{P}\}_{P \in \M}$ we calculate the OPE error. We aggregate across all $\{\pi^\ast_{P}\}_{P \in \M}$ by taking the average of the OPE errors and the worst-case, which can be seen in Figure \ref{fig:LQR_MML_vs_rest_OPO} (left).
\textbf{Metric:} We compute $|(J(\pi^\ast_{\widehat{P}},\widehat{P}) - J(\pi^{\ast}_{P^\ast},P^\ast))|,$ the OPO error.

\textbf{Note:} All calculations in LQR OPE/OPO are in expectation so no error bars need be included.

\textbf{Verifiability.} With the same setup as in OPE, now randomly sample 100k points in the interval $[-3,3] \times [-3,3]$, which is the support of the LQR system. We rerun the same experiment as in OPE except now we add $w \sim \mathcal{N}(0,\epsilon)$ noise to $V \in \V$ where $\epsilon \in \{0,.2,\ldots,.8,1\}$. We evaluate the error $|(J(\pi,\widehat{P}) - J(\pi,P^\ast))|$ over the 100k samples rather than in expectation as before. We run $5$ seeds and present the mean over the seeds with standard error. We smooth the resulting mean with a moving average filter of size 3. The result can be seen in Figure \ref{fig:LQR_verifiability}.

\subsubsection{Cartpole OPE}

Each $P \in \M$ takes the form $s' \sim \mathcal{N}(\mu(s,a), \sigma(s,a))$, where a NN outputs a mean, and logvariance representing a normal distribution around the next state. Each model has a two hidden layers and with $64$ units each and ReLU activation with final linear layer. We generate the behavior and target policy using a near-perfect DDQN-based policy $Q$ with a final softmax layer and adjustable parameter $\tau$: $\pi(a|s) \propto \exp(Q(s,a)/\tau).$ The behavior policy has $\tau = 1$, while the target policy has $\tau = 1.5$. We truncate all rollouts at $1000$ time steps and we calculate the true expected value using the monte-carlo average of $10000$ rollouts.

We model the class $\W\V$ as a RKHS as in Lemma \ref{lem:rkhs} with an RBF kernel. We do the same for VAML. The RKHS kernel we use for MML and VAML is given by $K(s,a,s') = K_1(s) K_2(a) K_3(s')$ and $K_3(s')$ respectively where $K_i$ are Gaussian Radial Basis Function (RBF) kernels with a bandwidth equal to the median of the pair-wise distances for each coordinate ($s,a,s'$ independently) over the batch.

For MML, we sample from $P$ a total of $5$ times and take the empirical mean to calculate the expectation over $P$ for the RKHS formula given in \ref{lem:rkhs}.

We run 20000 batches of size 128 and normalize the data over the batch. Our learning rate is $10^{-3}$ and we use Adam \citep[]{kingma2014Adam} optimizer. The estimate we use is the mean over the last 10 batches. We run $5$ random seeds per dataset size, and plot the log-relative MSE with standard error in Figure \ref{fig:OPE}.

\textbf{Note:} These hyperparameters remain the same across the different loss functions.

\textbf{Metric:} We compare the methods using the log-relative MSE metric: $\log(\frac{(J(\pi,\widehat{P}) - J(\pi,P^\ast))^2}{(J(\pi_b,P^\ast) - J(\pi,P^\ast))^2})$, which is negative when the OPE estimate $J(\pi, \widehat{P})$ is superior to the on-policy estimate $J(\pi_b, \widehat{P})$. The more negative, the better the estimate. To calculate $J(\pi,\widehat{P})$ we run $100$ trajectories in $\widehat{P}$ and take the mean.

\subsubsection{Inverted Pendulum OPO}

We generate the behavior data using a noisy feedback-linearized controller: $\pi_b(a|s)$ is uniformly random with probability $.3$ and is a feedback-linearized LQR controller (FLC)  with probability $.7$ where we use the FLC corresponding to LQR matrices $Q = 2 I_{2\times 2}, R = I_{2 \times 2}$. We truncate all rollouts at $200$ time steps. We fit $4$ feed-forward neural networks representing $P_1,\ldots,P_4$ where each is a deterministic model with two layers of $16$ weights and a Tanh activation followed with Linear. We use Adam \citep[]{kingma2014Adam} optimizer with $10^{-3}$ as the learning rate. Using different batches of size 64 on each $P_i$ and perform 5000 iterations for each model.

The RKHS kernel we use for MML and VAML is given by $K(s,a,s') = K_1(s) K_2(a) K_3(s')$ and $K_3(s')$ respectively where $K_i$ are Gaussian Radial Basis Function (RBF) kernels with a bandwidth equal to $1$.

For MML, we only sample from $P$ once to calculate the expectation over $P$ for the RKHS formula given in \ref{lem:rkhs}, since $P$ is deterministic.

Now we have $P(s'|s,a) = \frac{1}{4}\sum_{i=1}^4 P_i(s'|s,a)$. We calculate $\alpha = Median(\{ \|P_j(s,a) - s'\|_2 : j \in [1,\ldots,4], (s,a,s') \in X \subset D\})$ where $X$ is 10000 random samples from the dataset. We form an $\alpha$-USAD (see MOREL Section \ref{app:morel}) and construct a pessimistic MDP ($\tilde{P}, \tilde{R}$) (see Section \ref{app:morel}). We use PPO as our policy optimizer with the default settings from \cite[]{stablebaselines3}. We run PPO three times in the pessimistic MDP and take the policy that performs the best and report its performance. We keep track of the running maximum as we increase the dataset size. We plot the mean of the running maximums over the five seeds including standard error bars in Figure \ref{fig:OPO}.

\textbf{Note:} These hyperparameters remain the same across the different loss functions.

\textbf{Metric:} We look at the performance $J(\pi^\ast_{\widehat{P}},P^\ast)$ of a policy and compare it to $\pi^\ast$, learned from PPO. To calculate $J(\pi^\ast_{\widehat{P}},P^\ast)$ we run $100$ trajectories in $P^\ast$ and take the mean.

\subsection{MOREL}
\label{app:morel}

We give a brief explanation of MOREL \citep[]{kidambi2020morel} and its construction. The objective of MOREL is to make sure that the policy we learn does not take advantage of the errors in the simulator $P$. If there are errors in $P$ then a policy may think the agent can perform a particular state transition $(s,a,s')$ and $R(s',a')$ has high reward for some action $a'$. However, it's possible that such a transition $(s,a,s')$ may not occur in the true environment. Therefore, we modify our model $P(s'|s,a)$ in the following way:
$$ \tilde{P}(s'|s,a) = \begin{cases}
    \text{Terminate episode} \quad &U^\alpha(s,a) = 1 \\
    P(s'|s,a) &\quad \text{otherwise}
  \end{cases}
 $$
 where $U^\alpha(s,a) = 1$ if $\max_{i \in \{1,2,3,4\}}\|P_i(s'|s,a) - P(s'|s,a)\| \geq \alpha,$ otherwise $0$. In other words, we've modified the transition dynamics so that we do not trust our model $P$ unless all the $P_i$ are in agreement. We also modify our reward to be $$ \tilde{R}(s,a) = \begin{cases}
    \text{-100} \quad U^\alpha(s,a) = 1 \\
    R(s,a) \quad \text{otherwise}
  \end{cases}
 $$
 where $-100$ is chosen this value is well below any reward that the Inverted Pendulum environment generates.
Similarly, we penalize our policy for entering a state where we are uncertain. Together, this creates a pessimistic MDP.

\subsection{Additional Experiments}

\begin{figure}[!ht]
\minipage{.5\textwidth}
  \includegraphics[width=\linewidth]{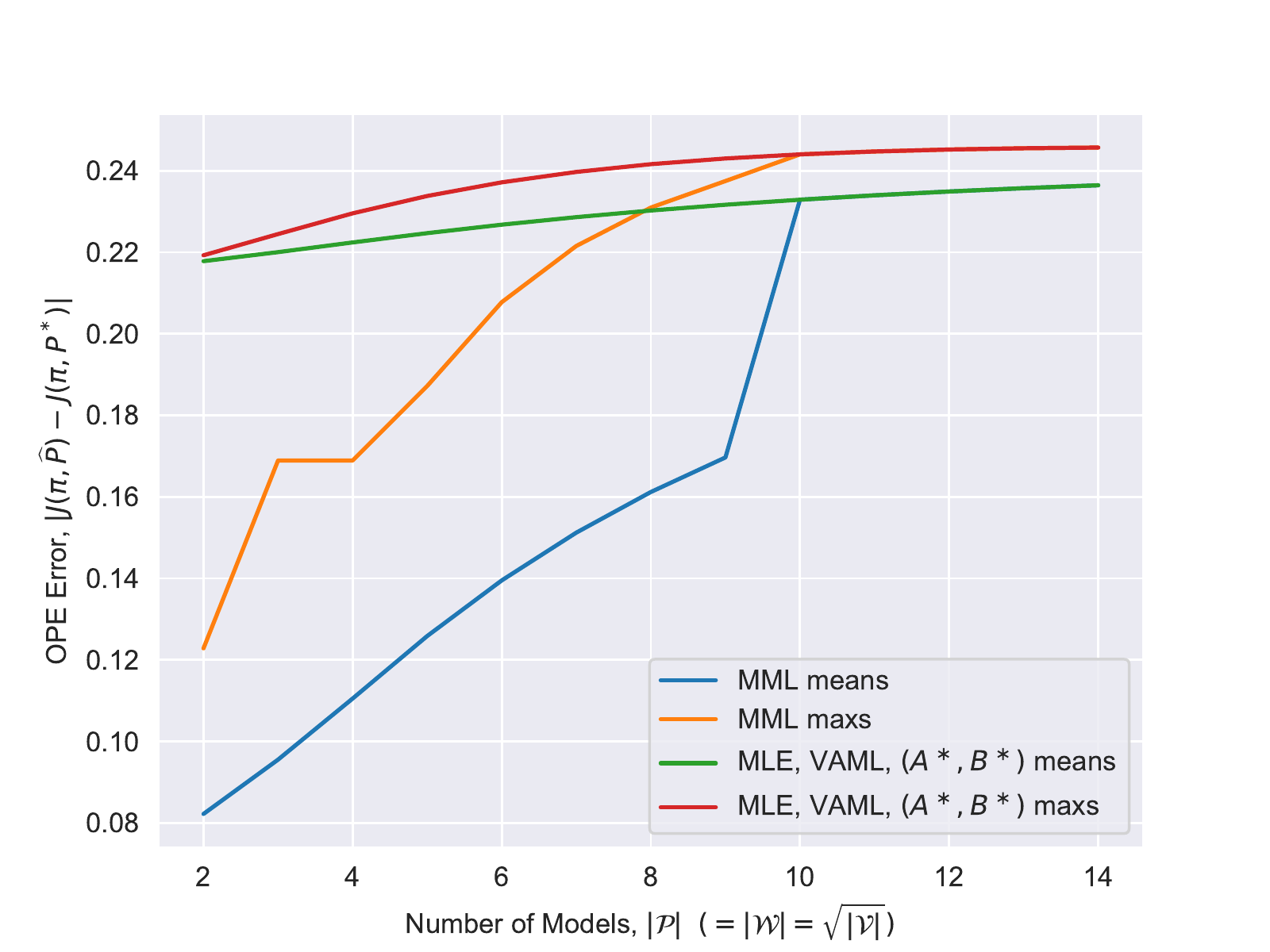}
\endminipage\hfill
\minipage{.5\textwidth}
  \includegraphics[width=\linewidth]{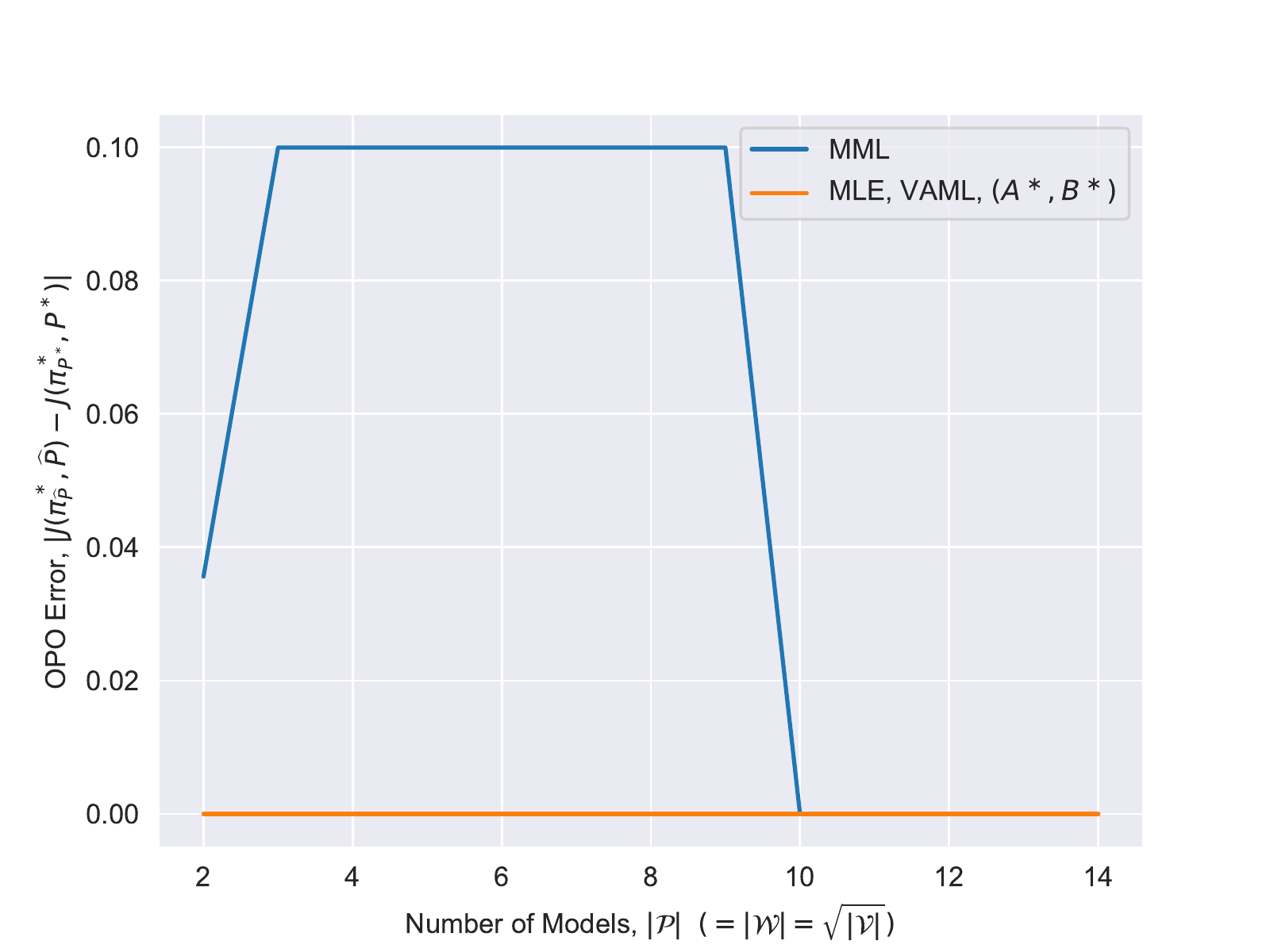}
\endminipage\hfill
\caption{\textit{ (LQR) As we increase $|W|,|V|$ then MML is forced to be robust to too many OPE problems and settles for the system $(A^\ast, B^\ast)$ since this is the only system robust to the most OPE problems.
}}
 \label{fig:LQR_MML_vs_rest_OPO}
 \vspace{-.1in}
\end{figure}

In the experiments for Figure \ref{fig:LQR_MML_vs_rest_OPO}, we consider what happens when we satisfy the realizability conditions for OPO. As we increase $|\M|$, we must also increase $|\W|, |\V|$ because each $P \in \M$ induces an optimal policy $\pi^\ast_P$ to which we have to make sure $w^{P^\ast}_{\pi^\ast_P} \in \W$ and $V^{P_i}_{\pi^\ast_P} \in \V$ for $\forall P_i \in \M$. In a sense, we are adding more OPE problems for MML to be robust to. In particular, we now have more policies $\{\pi^\ast_{P}\}_{P \in \M}$ to consider. As described earlier, for each $\pi \in \{\pi^\ast_{P}\}_{P \in \M}$ we calculate the OPE error. We aggregate across all $\{\pi^\ast_{P}\}_{P \in \M}$ by taking the average of the OPE errors and the worst-case, which can be seen in Figure \ref{fig:LQR_MML_vs_rest_OPO} (left). We plot the OPO error in Figure \ref{fig:LQR_MML_vs_rest_OPO} (right). What we see is that while $|\M|$ is small, MML is able to be robust to a certain number of OPE problems. But as we increase the number of OPE problems the average and max error increases until all methods select the same model, which is the OPO-optimal model, $(A^\ast, B^\ast)$.

\end{document}